\pdfoutput=1
\documentclass[11pt]{article}

\usepackage[ukrainian,vietnamese,english]{babel}
\usepackage{arabtex}
\usepackage{utf8}
\usepackage{amsmath}

\usepackage{emnlp22}
\usepackage{times}
\usepackage{latexsym}

\usepackage[T2A, T1]{fontenc}
\usepackage[utf8]{inputenc}

\usepackage{microtype}

\usepackage{inconsolata}

\usepackage{graphicx}
\usepackage{booktabs,multirow}
\usepackage{colortbl}

\definecolor{lightred}{rgb}{1, 0.70, 0.70}
\definecolor{lightorange}{rgb}{1, 0.90, 0.70}
\definecolor{lightgreen}{rgb}{0.7, 1, 0.7}
\definecolor{lightblue}{rgb}{0.7, 0.85, 1}

\definecolor{blueAA}{rgb}{0.99,0.99,1}
\definecolor{blueA}{rgb}{0.97,0.97,1}
\definecolor{blueB}{rgb}{0.87,0.87,1}
\definecolor{blueC}{rgb}{0.75,0.75,1}
\definecolor{blueF}{rgb}{0.35,0.35,1}

\definecolor{redA}{rgb}{1,0.95,0.95}
\definecolor{redB}{rgb}{1,0.85,0.85}
\definecolor{redC}{rgb}{1,0.7,0.7}
\definecolor{redD}{rgb}{1,0.55,0.55}
\definecolor{redE}{rgb}{1,0.45,0.45}
\definecolor{redF}{rgb}{1,0.35,0.35}

\usepackage{todonotes}

\title{DivEMT: Neural Machine Translation Post-Editing Effort \\ Across Typologically Diverse Languages}

\author{Gabriele Sarti{\normalfont{,}} Arianna Bisazza{\normalfont{,}} Ana Guerberof-Arenas{\normalfont{,}} Antonio Toral \vspace{2mm} \\
    Center for Language and Cognition (CLCG), University of Groningen \\
  \texttt{\{g.sarti, a.bisazza, a.guerberof.arenas, a.toral.ruiz\}@rug.nl}
}

\begin{document}
\maketitle

\begin{abstract}

We introduce DivEMT, the first publicly available post-editing study of Neural Machine Translation (NMT) over a typologically diverse set of target languages. Using a strictly controlled setup, 18 professional translators were instructed to translate or post-edit the same set of English documents into Arabic, Dutch, Italian, Turkish, Ukrainian, and Vietnamese. During the process, their edits, keystrokes, editing times and pauses were recorded, enabling an in-depth, cross-lingual evaluation of NMT quality and post-editing effectiveness. Using this new dataset, we assess the impact of two state-of-the-art NMT systems, Google Translate and the multilingual mBART-50 model, on translation productivity. We find that post-editing is consistently faster than translation from scratch. However, the magnitude of productivity gains varies widely across systems and languages, highlighting major disparities in post-editing effectiveness for languages at different degrees of typological relatedness to English, even when controlling for system architecture and training data size.
We publicly release the complete dataset\footnote{\href{https://github.com/gsarti/divemt}{\texttt{https://github.com/gsarti/divemt}} \\ \href{https://huggingface.co/datasets/GroNLP/divemt}{\texttt{https://huggingface.co/datasets/GroNLP/divemt}}} including all collected behavioral data, to foster new research on the translation capabilities of NMT systems for typologically diverse languages.
\end{abstract}

\section{Introduction}

Recent advances in neural language modeling and multilingual training have prompted a widespread adoption of machine translation (MT) technologies across an unprecedented range of world languages.
While the benefits of state-of-the-art MT for cross-lingual information access are undisputed~\citep{lommel_machine_2021},
its usefulness as an aid to professional translators varies considerably across domains, subjects and language combinations~\cite{zouhar-etal-2021-neural}. In the last decade, the MT community has been including an increasing number of languages in its automatic and human evaluation efforts~\citep{bojar-etal-2013-findings,wmt-2021-machine}.
However, the results of these evaluations are typically not directly comparable across different language pairs for various reasons.
First, reference-based automatic quality metrics are hardly comparable across different target languages~\citep{bugliarello-etal-2020-easier}.
Secondly, human judgments are collected independently for different language pairs, making their cross-lingual comparison vulnerable to confounding factors such as tested domains and training data sizes.
Similarly, recent work on NMT post-editing efficiency has focused on specific language pairs such as English-Czech~\citep{zouhar-etal-2021-neural}, German-Italian, German-French~\citep{laubli-etal-2019-post} and English-Hindi~\citep{ahsan-etal-2021-assessing}, but a controlled comparison across a set of typologically diverse languages is still lacking. 

\begin{figure}[t]
    \centering
    \includegraphics[width=\linewidth]{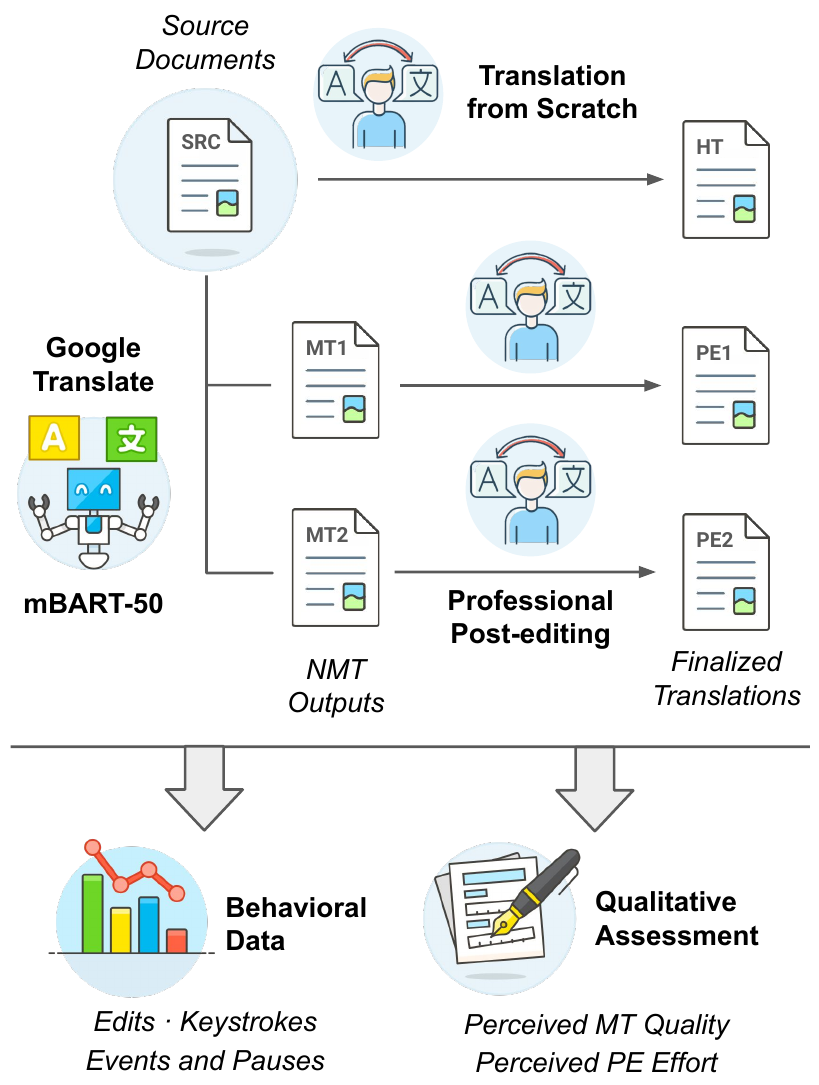}
  \caption{The DivEMT data collection process. For every English source document, 18 professional translators are tasked to translate it from scratch (HT) or post-edit NMT systems' outputs (PE$_1$/PE$_2$) into six typologically diverse target languages. Behavioral data and qualitative assessments are collected during and after the process respectively.}
  \label{fig:divemt-overview}
\end{figure}

In this work, we assess the usefulness of state-of-the-art NMT in professional translation with a strictly controlled cross-language setup (Figure~\ref{fig:divemt-overview}). Specifically, professionals were asked to translate the same English documents into six typologically different languages (Arabic, Dutch, Italian, Turkish, Ukrainian, and Vietnamese) using the same platform and guidelines.
Three \textit{translation modalities} were adopted: human translation from scratch (HT), post-editing of Google Translate's translation (PE$_1$), and post-editing of mBART-50's translation (PE$_2$), the latter being a state-of-the-art open-source, multilingual NMT system.
In addition to post-editing results, subjects' fine-grained editing behavior --- including keystrokes and time information --- was logged to measure productivity and effort across languages, systems and translation modalities. Finally, translators were asked to complete a qualitative assessment regarding their perceptions of MT quality and post-editing effort.
The resulting DivEMT dataset is to our best knowledge the first public resource allowing a direct comparison of professional translators' productivity and fine-grained editing information across a set of typologically-diverse languages.
\mbox{DivEMT} is publicly released alongside this paper as a unique resource to study the language- and system-dependent nature of NMT advances in real-world translation scenarios.

\section{Related Work}

\paragraph{Cross-lingual MT Evaluation} Before the advent of NMT,~\citet{birch-etal-2008-predicting}
studied how various language properties affected the quality of Statistical MT (SMT) across a sizeable sample of European language pairs.
The comparison, however, was solely based on BLEU, which is in fact not comparable across different target languages~\citep{bugliarello-etal-2020-easier}.
Recent work on neural models introduced %more principled measurements for the intrinsically challenging task of
more principled ways to measure the intrinsic difficulty of
language-modeling~\citep{gerz-etal-2018-relation,cotterell-etal-2018-languages,mielke-etal-2019-kind} and machine-translating~\citep{bugliarello-etal-2020-easier,bisazza-etal-2021-difficulty} different languages. 
However, achieving this reliably without any human evaluation remains an open research question.
Human evaluations of MT quality are routinely conducted during campaigns such as WMT~\citep{ws-2006-statistical,akhbardeh-etal-2021-findings} and IWSLT~\citep{cettolo-etal-2016-iwslt,cettolo-etal-2017-overview} among others, but their focus is on language- and domain-specific ranking of MT systems --- often leveraging non-professional annotators~\citep{freitag-etal-2021-experts} --- rather than cross-lingual quality comparisons.
Concurrently to this work,  \newcite{licht22-consistent} proposed a new human evaluation protocol to improve consistency in cross-lingual MT quality assessment.

\paragraph{Post-editing NMT} Measuring post-editing effort across its \textit{temporal, cognitive, and technical} dimensions~\citep{krings_repairing_2001} is a well-established way to assess the effectiveness and efficiency of MT as a component of specialized translation workflows. 
Seminal post-editing studies highlighted an increase in translators' productivity following MT adoption~\citep{guerberof-2009-productivity,green_efficacy_2013, laubli-etal-2013-assessing, plitt_productivity_2010, parra_escartin_machine_2015}. 
However, they also struggled to identify generalizable findings due to confounding factors like output quality, content domains, and high variance across language pairs and human subjects.
With the advent of NMT, productivity gains of the new approach were extensively compared to those of  SMT, the highly-customized dominant paradigm at the time~\citep{castilho_is_2017, bentivogli-etal-2016-neural, toral_post-editing_2018, laubli-etal-2019-post}. Initial results were promising for NMT due to its better fluency and overall results.
Moreover, translators were shown to prefer NMT over SMT for post-editing, although a pronounced productivity increase was not always present.
More recent work highlighted the productivity gains driven by NMT post-editing in a wider array of languages that were previously challenging for MT,
such as English-Dutch~\citep{daems_translation_2017}, English-Hindi~\citep{ahsan-etal-2021-assessing}, 
English-Greek~\citep{stasimioti-sosoni-2020-translation}, 
English-Finnish and English-Swedish~\citep{koponen-etal-2020-mt},
all showing a considerable variance among language pairs and subjects.
Interestingly,~\citet{zouhar-etal-2021-neural} found NMT post-editing speed to be comparable to translation from scratch in English-Czech, and highlighted a disconnect between moderate increases in automatic MT quality metrics and better post-editing productivity.
In sum, research on post-editing NMT generally reports increased fluency and output quality, but productivity gains are hardly generalizable across language pairs and domains.
%\AG{Therefore, it is necessary to look at specific language combinations, especially those where less data is available, and to consider the aforementioned factors as they might impact the final results on effort.}
Importantly, to our knowledge, no previous work has studied NMT post-editing over a set of typologically different languages while controlling for the effects of content types and domains, NMT engines, and translation interfaces.

\section{The DivEMT Dataset}

DivEMT's main purpose is to assess the usefulness of state-of-the-art NMT for professional translators and to study how this usefulness varies across target languages with different typological properties.
We present  below our data collection setup, which strikes a balance between simulating a realistic professional translation workflow and maximizing the comparability of results across languages. 

\subsection{Subjects and Task Scheduling} 
To control for the effect of individual translators' preferences and styles, we involve a total of 18 subjects (three per target language). During the experiment, each subject receives a series of short \textit{documents} (3 to 5 sentences each) where the source text is presented in isolation (HT) or alongside a translation proposal produced by one of the NMT systems (PE$_1$, PE$_2$). The experiment comprises two phases: 
During the \textbf{warm-up phase} a set of 5 documents is translated by all subjects following the same, randomly sampled sequence of modalities (HT, PE$_1$ or PE$_2$). This phase allows the subjects to get used to the setup and enables us to spot possible issues in the logged behavioral data before moving forward.\footnote{Warm-up data are excluded from the analysis of Section~\ref{sec:pe_effort}.}
In the \textbf{main collection phase}, each subject is asked to translate documents in a pseudo-random sequence of modalities. This time, however, the sequence is different for each translator and chosen so that each document gets translated in all three modalities. 
This allows us to measure translation productivity independently from the subject's productivity and document-specific difficulties.
A graphical overview of this process is shown in Figure~\ref{fig:divemt-overview}, with additional  details given in Appendix~\ref{app:modality-scheduling}.
As productivity and other behavioral metrics can only be estimated with a sizable sample, we prioritize the number of documents over the number of subjects per language during budget allocation. A larger set of post-edited documents also provides more insight in the error type distribution of NMT systems across different language pairs, an analysis which we leave to future work.

All subjects are professional translators with at least 3 years of professional experience, at least one year of post-editing experience and strong proficiency with CAT tools.\footnote{Additional subjects' details are available in Appendix~\ref{app:subject-info}.}
Translators were provided with links to the source articles to facilitate contextualization, were asked to produce translations of publishable quality and were instructed not to use any external MT engine to produce their translations. Assessing the final quality of the post-edited material is out of the scope of the current study, although we realize that this is an important consideration to assess usability in a professional context. A summary of our translation guidelines is provided in Appendix~\ref{app:guidelines}.

\subsection{Choice of Source Texts}

The selected documents represent a subset of the FLORES-101 benchmark~\citep{FLORES101} consisting of sentences taken from English Wikipedia, and covering a mix of topics and domains.\footnote{We use a balanced sample of articles sourced from WikiNews, WikiVoyage and WikiBooks.} While professional translators generally specialize in one or a few domains, we opt for a mix-domain dataset to minimize domain adaptation efforts by the subjects and maximize the generalizability of our results. %\AG{I do not think this last sentence is strictly true. I would say "This meant that translators had to consult a variety of sources to process the translations and we expected the learning curve not be as steep as when translating the same domain or topic, i.e. that translators would process segments faster as they progress."}
%AB: To clarify for Ana, we (MT'ers) think that working in only  one specific domain with MT's that are not domain-tuned could have reflected a very specific (=non generalizable) weakness of the system as opposed to a general language pattern. In any case we have rephrased the sentence.
%
Importantly, FLORES-101 includes high-quality human translations into 101 languages, which makes it possible to automatically estimate NMT quality and discard excessively low-scoring models or language pairs before our experiment.
FLORES-101 also provides useful metadata, e.g. source URL, which allows us to ensure the absence of public translations of the selected contents, which could be leveraged by translators and compromise the validity of our setup.
The documents used for our study are fragments of contiguous sentences extracted from Wikipedia articles that compose the original FLORES-101 corpus. Even if small, the context provided by document structure allows us to simulate a more realistic translation workflow if compared to out-of-context sentences.

Based on our available budget, we select 112 English documents from the \textit{devtest} portion of FLORES-101 corresponding to 450 sentences and 9626 words. More details on the data selection process are provided in Appendix~\ref{app:doc-select}.

\begin{table}\centering\small
\begin{tabular}{@{} l @{\ \ \ } l @{\ \ \ } c @{\ \ } c @{\ \ \ } c @{\ \ \ } c @{\ \ \ } c @{}}
\toprule 
& \textbf{Genus:Family} & \textbf{\textit{d}}$_{syn}$ & \textbf{Morph.} & \textbf{MSP} &  \textbf{TTR} & \textbf{Script}\\
\midrule
\textsc{Eng} & IE:Germanic & -- & Fus & \cellcolor{redB}1.17 & \cellcolor{redC} 0.28 & latin \\ 
\midrule
\textsc{Ara} & \cellcolor{blueC}Af:Semitic & \cellcolor{blueB} 0.57  & Ifx & \cellcolor{redD}1.67 & \cellcolor{redE} 0.46  &  arab. \\ 
\textsc{Nld} & \cellcolor{blueA}IE:Germanic & \cellcolor{blueAA} 0.49 & Fus & \cellcolor{redB}1.16 & \cellcolor{redC} 0.28 & latin \\ 
\textsc{Ita} & \cellcolor{blueB}IE:Romance & \cellcolor{blueA} 0.51 & Fus & \cellcolor{redC}1.30 & \cellcolor{redC}  0.30 & latin \\ 
\textsc{Tur} & \cellcolor{blueC}Alt:Turkic & \cellcolor{blueC} 0.70 & Agg & \cellcolor{redF}2.28 & \cellcolor{redE} 0.50 & latin  \\ 
\textsc{Ukr} & \cellcolor{blueB}IE:Slavic & \cellcolor{blueA} 0.51 & Fus & \cellcolor{redC}1.42  & \cellcolor{redE} 0.47 & cyril. \\ 
\textsc{Vie} & \cellcolor{blueC}Au:VietMuong & \cellcolor{blueB} 0.57 & Iso & \cellcolor{redA}1.00 & \cellcolor{redA} 0.12 & latin \\ 
\bottomrule
\end{tabular}
 \caption{Typological diversity of our language sample.  \textbf{IE}: Indo-European, \textbf{Af}: Afro-Asiatic, \textbf{Alt}: Altaic, \textbf{Au}: Austro-Asiatic. 
 \textbf{\textit{d}}$_{syn}$: Syntactic distance w.r.t. English~\citep{lin-etal-2019-choosing}.
 \textbf{Fus}: fusional, \textbf{Ifx}: introflexive, \textbf{Agg}: agglutinative, \textbf{Iso}: isolating.
 \textbf{MSP}: Mean size of paradigm, from~\citet{coltekin22-morph-complexity}.
 \textbf{TTR}: Type-token ratio measured on FLORES-101. Shading indicates \textcolor{blueF}{genetic/syntactic relatedness to English} and \textcolor{redF}{morphological complexity/lexical richness}.
}
 \label{tab:languages-small}
\end{table}

\subsection{Choice of Languages}
\label{sec:choice-languages}
Training data is among the most important factors in defining the quality of a NMT system.
Unfortunately, using strictly comparable or multi-parallel datasets, like Europarl~\citep{koehn-2005-europarl} or the Bible  corpus \citep{mayer-cysouw-2014-creating}, would dramatically restrict the diversity of languages available to our study, or imply a prohibitively low translation quality on general-domain text. 
In order to minimize the effect of training data disparity while maximizing language diversity, we choose representatives of six different language families for which comparable amounts of training data are available in our open-source model, namely \textbf{Arabic}, \textbf{Dutch}, \textbf{Italian}, \textbf{Turkish}, \textbf{Ukrainian}, and \textbf{Vietnamese}. As shown in Table~\ref{tab:languages-small}, our language sample ensures a good diversity in terms of language family and relatedness to English, type of morphological system, morphological complexity --- measured by  mean size  of paradigm (MSP,~\citeauthor{xanthos-etal-2011-role}~\citeyear{xanthos-etal-2011-role}) --- and script. We also report type-token ratio (TTR), the only language property that was found to correlate significantly with translation difficulty in a sample of European languages~\citep{bugliarello-etal-2020-easier}.
While the amount of language-specific parallel sentence pairs used for the multilingual fine-tuning of mBART-50 varies widely (4K$<$\textit{N}$<$45M), all our selected language pairs fall within the 100K-250K range (mid-resourced, see Table~\ref{tab:flores-perf}), enabling a fair cross-lingual performance comparison.

\begin{table}\centering\small
\begin{tabular}{l c | c c }%c}
\toprule & \textbf{GTrans (PE$_1$)} & \textbf{mBART-50 (PE$_2$)} & \textbf{\# Pairs} \\%& Source \\
\midrule
\textsc{Ara} & \textbf{34.1} / \textbf{65.6} / \textbf{.737} & 17.0 / 48.5 / .452 & 226K \\% IWSLT'17 \\ 
\textsc{Nld} & \textbf{29.1} / \textbf{60.0} / \textbf{.667} & 22.6 / 53.9 / .532 & 226K \\%& IWSLT'17.mlt \\ 
\textsc{Ita} & \textbf{32.8} / \textbf{61.4} / \textbf{.781} & 24.4 / 54.7 / .648 & 233K \\%& IWSLT'17.mlt  \\ 
\textsc{Tur} & \textbf{35.0} / \textbf{65.5} / \textbf{1.00} & 18.8 / 52.7 / .755 & 204K \\%& WMT'17 \\ 
\textsc{Ukr} & \textbf{31.1} / \textbf{59.8} / \textbf{.758} & 21.9 / 50.7 / .587 & 104K \\%& TED58 \\ 
\textsc{Vie} & \textbf{45.1} / \textbf{61.9} / \textbf{.724} & 34.7 / 54.0 / .608 & 127K \\%& IWSLT'15 \\ 
\bottomrule
\end{tabular}
  \caption{MT quality of the selected NMT systems for English-to-Target translation on the full FLORES-101 devtest split, in \textsc{Bleu / ChrF / Comet} format. Best scores are highlighted in \textbf{bold}. We report the number of sentence pairs used for mBART-50 multilingual fine-tuning by~\citet{tang-etal-2021-multilingual}.}
  \label{tab:flores-perf}
\end{table}

\begin{table*}
    \centering
    \footnotesize
    \begin{tabular}{p{0.15in} p{0.15in} p{\textwidth - 0.75in}}
        \toprule
        \textsc{Eng} & \textsc{Src} & Inland waterways can be a good theme to base a holiday around. \\
        \midrule
        \textsc{Ara} & HT & \setcode{utf8}{\< يمكن أن تكون الممرات المائية الداخلية خياراً جيدًا لتخطيط عطلة حولها. >}\\
        & MT & \includegraphics[height=4.5mm]{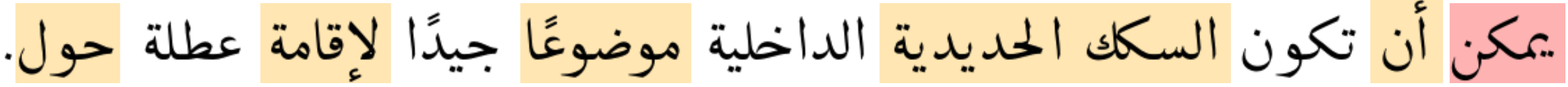} \\%\< يمكن أن تكون السكك الحديدية الداخلية موضوعًا جيدًا لإقامة عطلة حول. >\\
        & PE & \includegraphics[height=4.5mm]{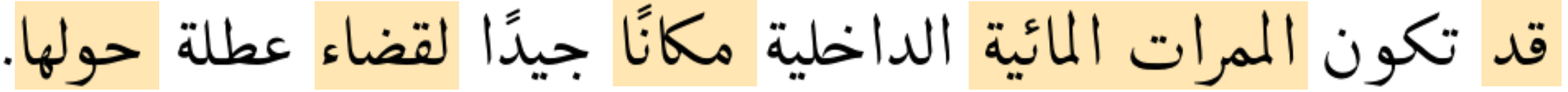} \\%\< قد تكون الممرات المائية الداخلية مكانًا جيدًا لقضاء عطلة حولها.>\\
        \midrule
        \textsc{Nld} & HT & Binnenlandse waterwegen kunnen een goed thema zijn voor een vakantie. \\
        & MT & Binnenwaterwegen kunnen een goed thema zijn om een vakantie rond te \colorbox{lightorange}{zetten}. \\
        & PE & Binnenwaterwegen kunnen een goed thema zijn om een vakantie rond te \colorbox{lightorange}{organiseren}. \\
        \midrule
        \textsc{Ita} & HT & I corsi d'acqua dell'entroterra possono essere un ottimo punto di partenza da cui organizzare una vacanza. \\
        & MT & I corsi d'acqua interni possono essere un buon tema \colorbox{lightorange}{per fondare} una vacanza. \\
        & PE & I corsi d'acqua interni possono essere un buon tema \colorbox{lightgreen}{su} \colorbox{lightorange}{cui basare} una vacanza. \\
        \midrule
        \textsc{Tur} & HT & İç bölgelerdeki su yolları, tatil planı için iyi bir tema olabilir. \\
        & MT & İç \colorbox{lightorange}{suyolları,} tatil için uygun bir tema olabilir.\\
        & PE & İç \colorbox{lightorange}{sular} tatil için uygun bir tema olabilir.\\
        \midrule
        \textsc{Ukr} & HT & \foreignlanguage{ukrainian}{Можна спланувати вихідні, взявши за основу подорож внутрішніми водними шляхами.} \\
        & MT & \foreignlanguage{ukrainian}{\colorbox{lightorange}{Водні шляхи можуть} \colorbox{lightblue}{бути} \colorbox{lightorange}{хорошим об 'єктом} \colorbox{lightblue}{для} \colorbox{lightorange}{базування відпочинку навколо}.} \\
        & PE & \foreignlanguage{ukrainian}{\colorbox{lightgreen}{Місцевість} \colorbox{lightorange}{навколо внутрішніх водних шляхів може} \colorbox{lightblue}{бути} \colorbox{lightorange}{гарним вибором} \colorbox{lightblue}{для} \colorbox{lightorange}{органiзацiї відпочинку.}} \\
        \midrule
        \textsc{Vie} & HT & \foreignlanguage{vietnamese}{Du lịch trên sông có thể là một lựa chọn phù hợp cho kỳ nghỉ.} \\
        & MT & \foreignlanguage{vietnamese}{\colorbox{lightorange}{Các tuyến nước} \colorbox{lightblue}{nội địa} \colorbox{lightorange}{có thể} là một \colorbox{lightred}{chủ đề tốt} \colorbox{lightorange}{để xây dựng một kì nghỉ.}} \\
        & PE & \foreignlanguage{vietnamese}{\colorbox{lightorange}{Du lịch bằng đường thủy} \colorbox{lightblue}{nội địa} là một \colorbox{lightorange}{ý tưởng nghỉ dưỡng không tồi.}} \\
        \bottomrule
    \end{tabular}
    \caption{A DivEMT corpus entry, including the English source (\textsc{Src}), its translation from scratch (HT), the MT output of mBART-50 (MT) and its post-edited version (PE) for all languages. We highlight \colorbox{lightgreen}{insertions}, \colorbox{lightred}{deletions}, \colorbox{lightorange}{substitutions} and \colorbox{lightblue}{shifts} computed with Tercom~\cite{snover-etal-2006-study}. Full examples available in Appendix~\ref{app:examples}.}
    \label{tab:data_example_small}
\end{table*}

\subsection{Choice of MT Systems}
\label{sec:choice-systems}
While most of the best-performing general-domain NMT systems are commercial, experiments based on such systems are not replicable as their back-ends get silently updated over time. Moreover, without knowing the exact training specifics, we cannot attribute differences in the cross-lingual results to intrinsic language properties.
We balance these observations by including two NMT systems in our study: \textbf{Google Translate} (GTrans)\footnote{Evaluation performed in October 2021.} as a representative of commercial quality, and \textbf{mBART-50 one-to-Many}\footnote{\href{https://huggingface.co/facebook/mbart-large-50-one-to-many-mmt}{\texttt{facebook/mbart-large-50-one-to-many}}}~\citep{tang-etal-2021-multilingual} as a representative of state-of-the-art open-source multilingual NMT technology. The original multilingual BART model~\citep{liu-etal-2020-multilingual-denoising} is an encoder-decoder transformer model pre-trained on monolingual documents in 25 languages.~\citet{tang-etal-2021-multilingual} extend mBART by further pre-training on 25 new languages and performing \textit{multilingual translation fine-tuning} for the full set of 50 languages, producing three configurations of multilingual NMT models: many-to-one, one-to-many, and many-to-many.
Our choice of mBART-50 is largely motivated by its maneageable size, its good performances across the set of evaluated languages (see Table~\ref{tab:flores-perf}) and its adoption for other NMT~\citep{liu-etal-2021-continual} and post-editing~\citep{fomicheva-etal-2020-mlqepe} studies. 
Although mBART-50 performances are usually comparable or slightly worse than the ones of tested bilingual NMT models,\footnote{See Appendix~\ref{app:other} for automatic MT quality results by five different models over a larger set of 10 target languages.} using a multilingual model allows us to evaluate the downstream effectiveness of a single, unified system trained on pairs evenly distributed across tested languages. Finally, adopting two systems with marked differences in automatic evaluation scores allows us to estimate how a significant increase in metrics such as \textsc{Bleu}, \textsc{ChrF} and \textsc{Comet}~\citep{papineni-etal-2002-bleu,popovic-2015-chrf,rei-etal-2020-comet} impacts downstream productivity across languages in a realistic post-editing scenario.

\subsection{Translation Platform and Collected Data} 

Translators were asked to use \textsc{Pet}~\citep{aziz-etal-2012-pet}, a computer-assisted translation tool that supports both translating from scratch and post-editing.
This tool was chosen because (i) it logs information about the post-editing process, which we use to assess effort (see Section~\ref{sec:pe_effort}); and (ii) it is a mature research-oriented tool that has been successfully used in several previous studies~\citep{koponen2012post,toral_post-editing_2018}. 
The minimalistic nature of \textsc{Pet} interface and functionalities limits its application in commercial translation activities, making it generally unfamiliar for professional translators. We consider this aspect an advantage in light of our controlled setup since it allows us to avoid additional confounding effects or disparities stemming from tools-specific capabilities and different degrees of proficiency with the software.
We also observe that, due to the varied and generic nature of the selected documents, functionalities such as concordance and translation memory matches would have proven much less useful in our setup. We collect three types of data:

\begin{itemize}
    \setlength{\itemsep}{0em}
    \setlength{\itemindent}{0em}
    \item \textbf{Resulting translations} produced by translators in either HT or PE modes, constituting a multilingual corpus with one source text and 18 translations (one per language-modality combination) exemplified in Table~\ref{tab:data_example_small}.

    \item \textbf{Behavioral data} for translated sentences, including editing time, amount and type of keystrokes (content, navigation, erase, etc.), and number and duration of pauses above 300/1000 milliseconds~\citep{lacruz-etal-2014-cognitive}.

    \item \textbf{Pre- and post-task questionnaire}. The former focuses on demographics, education, and work experience with translation and post-editing. The latter elicits subjective assessments of post-editing quality, effort and enjoyability compared to translating from scratch.
\end{itemize}

\section{Post-Editing Effort Across Languages}\label{sec:pe_effort}

\begin{figure}[t]
    \centering
    \includegraphics[width=\linewidth]{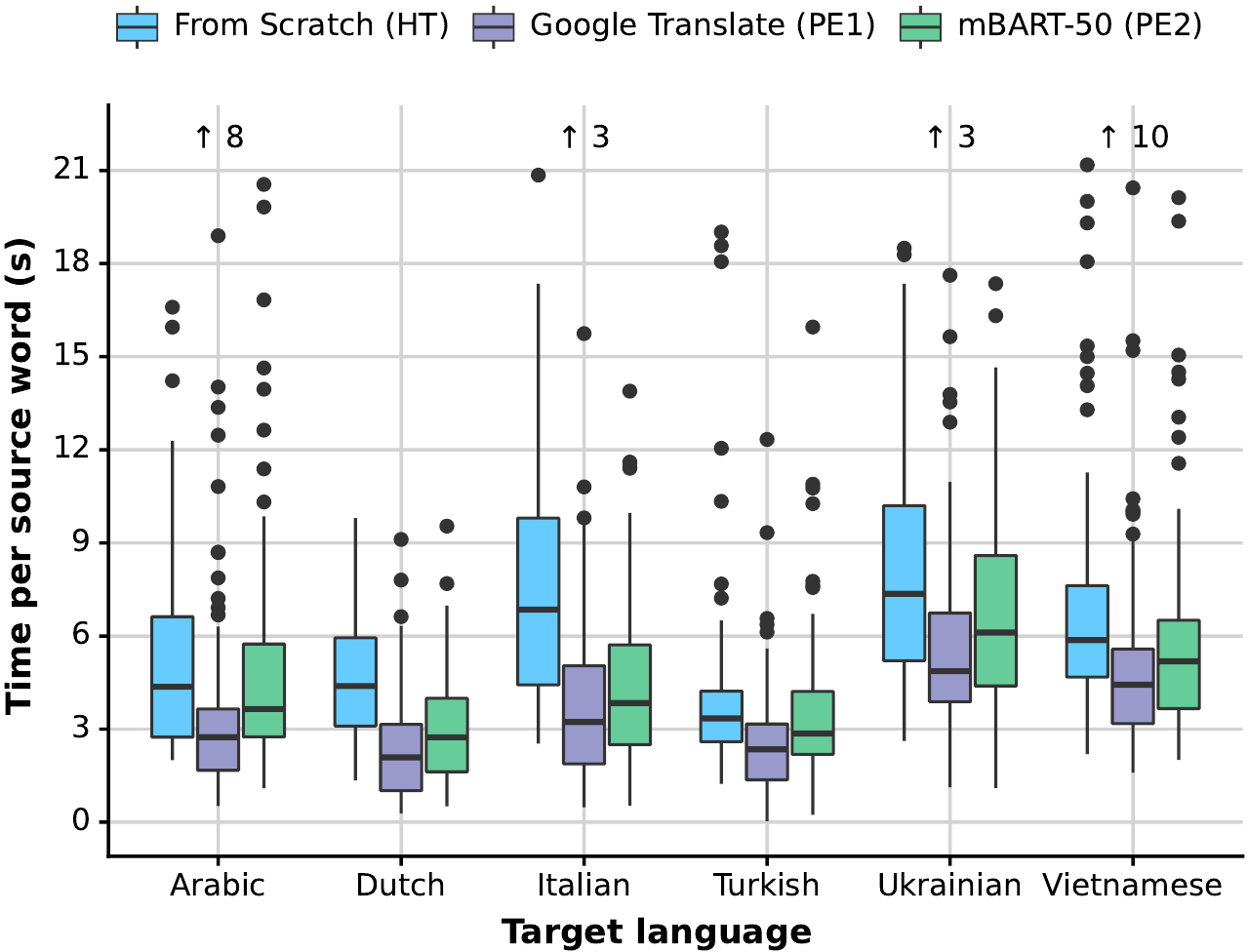}
  \caption{Temporal effort across languages and translation modalities, measured in seconds per processed source word. Each point represents a document, with higher scores denoting slower editing. $\uparrow$: amount of data points per language not shown in the plot.
  }
  \label{fig:time-per-src-word}
\end{figure}

In this section, we use the DivEMT dataset to quantify the post-editing effort of professional translators across our diverse set of target languages.
We consider two main objective indicators of editing effort, namely \textit{temporal measurements} (and related productivity gains) and
\textit{post-editing rates}, measured by the Human-targeted Translation Edit Rate (HTER,~\citeauthor{snover-etal-2006-study}~\citeyear{snover-etal-2006-study}).
Finally, we assess the subjective perception of PE gains by examining the post-task questionnaires. We reiterate that all scores in this section are computed on the same set of source sentences for all languages, resulting in a faithful cross-lingual comparison of post-editing effort thanks to DivEMT's controlled setup.

\subsection{Temporal Effort and Productivity Gains}

We start by comparing \textit{task time} (seconds per processed source word) across languages and modalities. For this purpose, edit times are computed for every document in every language without considering the presence of multiple translators for every language.
As shown in Figure~\ref{fig:time-per-src-word}, translation time varies considerably across languages even when no MT system is involved (HT), suggesting an intrinsic variability in translation complexity for different subjects and language pairs. 
Indeed, for the HT modality, the time required for the `slowest' target languages (Italian, Ukrainian) is roughly double the `fastest' one (Turkish). This pattern cannot be easily explained and contrasts with factors commonly tied to MT complexity, such as source-target morphological richness and language relatedness~\citep{birch-etal-2008-predicting,belinkov-etal-2017-neural}.
On the other hand, we find the relation PE$_1$ > PE$_2$ > HT (PE$_1$ fastest, PE$_2$ medium speed, HT slowest) to hold for all the evaluated languages.

\begin{table}\centering\small
\begin{tabular}{c c c c | r r }
\toprule
     & \multicolumn{3}{c}{\textbf{\textsc{Prod} $\uparrow$}} & \multicolumn{2}{c}{\textbf{$\Delta$HT $\uparrow$}} \\
     \cmidrule(lr){2-4}
     \cmidrule(lr){5-6}
     & HT & PE$_1$ & PE$_2$ & PE$_1$ & PE$_2$ \\
\midrule
\textsc{Ara} & 13.1 & 21.7 & 16.3 & +84\%  & +10\%  \\
\textsc{Nld} & 13.6 & 28.7 & 21.7 & +119\% & +61\%  \\
\textsc{Ita} & 8.8  & 18.6 & 15.6 & +96\%  & +95\%  \\
\textsc{Tur} & 17.9 & 25.5 & 21.0 & +34\%  & +12\%  \\
\textsc{Ukr} & 8.0  & 12.3 & 9.8  & +71\%  & +14\%  \\
\textsc{Vie} & 10.2 & 13.0 & 11.1 & +32\%  & +23\%  \\
\bottomrule
\end{tabular}
  \caption{Median productivity (\textsc{Prod}, \# processed source words per minute) and median \% post-editing speedup ($\Delta$HT) for all analyzed languages and modalities. Arrows denote the direction of improvement.}
  \label{tab:productivity}
\end{table}

For a measure of productivity gains that is easier to interpret and more in line with translation industry practices, we turn to \textit{productivity} expressed in source words processed per minute and compute the \textit{speed-up} induced by the two post-editing modalities over translating from scratch ($\Delta$HT). Table~\ref{tab:productivity} presents our results.
\textbf{Across systems}, we find that \emph{large} differences among automatic MT quality metrics indeed reflect on post-editing effort, suggesting a nuanced picture that is complementary to the findings of~\citet{zouhar-etal-2021-neural}. While post-editing time gains were observed to quickly saturate for slight changes in high-quality MT, we find that moving from medium-quality to high-quality MT yields meaningful productivity improvements across most evaluated languages.
\textbf{Across languages}, too, the magnitude of productivity gains ranges widely, from doubling in some languages (Dutch PE$_1$, Italian PE$_1$ and PE$_2$) to only about 10\% (Arabic, Turkish and Ukrainian PE$_2$).
When only considering the better performing system (PE$_1$), post-editing remains clearly beneficial in all languages despite the high variability in $\Delta$HT scores.
Results are more nuanced for the open-source system (PE$_2$), with three out of six languages displaying only marginal gains (<15\% in Arabic, Turkish and Ukrainian). 
Despite its overall lower performance, mBART-50 (PE$_2$) is the only system enabling a fair comparison across languages (from the point of view of training data size and architecture, see Section~\ref{sec:choice-systems}). Interestingly, if we focus on the gains induced by this system, factors like language relatedness and morphological complexity become relevant. Specifically, Italian (+95\%), Dutch (+61\%) and Ukrainian (+14\%) are genetically and syntactically related to English, but Ukrainian has a richer morphology (see Table~\ref{tab:languages-small}).
On the other hand, Vietnamese (+23\%), Turkish (+12\%) and Arabic (+10\%) all belong to different families. However, Vietnamese is isolating (little to no morphology), while Turkish and Arabic have very rich morphological systems (resp. agglutinative and introflexive, the latter of which is especially problematic for subword segmentation,~\citeauthor{amrhein-sennrich-2021-suitable-subword}~\citeyear{amrhein-sennrich-2021-suitable-subword}).
Other differences are however harder to explain. For instance, Dutch is closely related to English and has a simpler morphology than Italian, but its productivity gain with mBART-50 is lower (61\% vs 95\%). This finding is accompanied by an important gap in \textsc{Bleu} and \textsc{Comet} scores achieved by mBART-50 on the two languages (22.6 vs 24.4 \textsc{Bleu} and 0.532 vs 0.648 \textsc{Comet} for Dutch vs Italian, resp.) which cannot be explained by training data size. 

In summary, our findings confirm the overall positive impact of NMT post-editing on translation productivity observed in previous PE studies. However, we note how \textbf{the magnitude of this impact is highly variable across systems and languages}, with inter-subject variability also playing an important role, in line with previous studies~\citep{koponen-etal-2020-mt} (see Section~\ref{sec:limitations} for more details).
The small size of our language sample does not allow us to draw direct causal links between specific typological properties and post-editing efficiency.
That said, we believe these results have important implications on the claimed `universality' of current state-of-the-art MT and NLP systems, mostly based on the Transformer architecture~\citep{transformer_2017} and BPE-style subword segmentation techniques~\citep{sennrich-etal-2016-neural}.

\subsubsection{Modeling Temporal Effort}

\begin{table}
    \small
    \centering
    \begin{tabular}{lrrr}
        \toprule
         \textbf{Predictor} & \textbf{Estim.} & \textbf{p-value} & \textbf{Sig.}\\
        \midrule
        \texttt{(intercept)}
          & 4.92 & 1.12e-11&*** \\ 
        \texttt{source length}
          & 0.38 & < 2e-16&***\\
          \midrule
        \texttt{lang\_ara} &            -0.49 &0.1209&\\
        \texttt{lang\_ita} &            -0.14 &0.6407&\\
        \texttt{lang\_nld} &            -0.58 &0.0733& x\\
        \texttt{lang\_tur} &            -0.82 &0.0162& *\\
        \texttt{lang\_vie} &            -0.24 &0.4254&\\
        \midrule
        \texttt{task\_pe1} &            -0.49 &< 2e-16&***\\
        \texttt{task\_pe2} &            -0.22 & 1.77e-07&***\\
        \midrule
        \texttt{lang\_ara:task\_pe1}&   -0.11 &0.0505&x\\
        \texttt{lang\_ita:task\_pe1}&   -0.40 &8.97e-12&***\\
        \texttt{lang\_nld:task\_pe1}&   -0.41 &5.74e-12&***\\
        \texttt{lang\_tur:task\_pe1}&   -0.14 &0.0194&*\\
        \texttt{lang\_vie:task\_pe1}&    0.13 &0.0290&*\\
        \midrule
        \texttt{lang\_ara:task\_pe2}&    0.05 &0.3535&\\
        \texttt{lang\_ita:task\_pe2}&   -0.39 &3.30e-11&***\\
        \texttt{lang\_nld:task\_pe2}&   -0.29 &4.46e-07&***\\
        \texttt{lang\_tur:task\_pe2}&    0.03 &0.5811&\\
        \texttt{lang\_vie:task\_pe2}&    0.04 &0.5289&\\
        \bottomrule
    \end{tabular}
    \caption{LMER modeling results using translation time as the dependent variable. The reference levels for predictors \texttt{lang} and \texttt{task} are Ukrainian and Translation from scratch (HT), respectively. Estimate impact on edit time for every predictor is provided in log seconds. Significance: *** = < 0.001,  * = < 0.05, x = < 0.1}
    \label{tab:significance}
\end{table}

Given the high variability among translators, segments and translation modalities, we assess the validity of our observations via statistical analysis of temporal effort using a linear mixed-effects regression model (LMER,~\citeauthor{lindstrom-bates-1988-lmer}~\citeyear{lindstrom-bates-1988-lmer}), following~\citet{green_efficacy_2013} and~\citet{toral_post-editing_2018}. We fit our model on $n=7434$ instances, corresponding to 413 sentences translated by 18 translators\footnote{Outliers were removed beforehand, see Appendix D.}, using translation time as the dependent variable.
Our fixed predictors include translation modality, target language, their interaction and length of source segment in characters.\footnote{The document processing order was originally included to identify possible longitudinal effects but was removed due to a lack of significant improvements.}
Our random effects structure includes random intercepts for different segments (nested with documents) and translators, as well as a random slope for modality over individual segments.\footnote{Additional modeling details available in Appendix~\ref{app:modeling}.} Table~\ref{tab:significance} presents the set of predictors included in the final model, an estimate of their impact on edit times and their significance.
We find both PE modalities to significantly reduce translation times ($p<0.001$), with PE$_1$ being significantly faster than PE$_2$ ($p<0.001$) across all languages.
Taking the language for which HT is slowest (Ukrainian) as the reference level, 
the reduction in time brought by Google is significantly more pronounced for Italian, Dutch ($p<0.001$), and Turkish ($p<0.05$).
For mBART-50, however, we only observe significantly more pronounced increases in productivity for Italian and Dutch ($p<0.001$) compared to the reference. We find these results to corroborate the observations of the previous section.

\begin{figure}
    \centering
    \includegraphics[width=\linewidth]{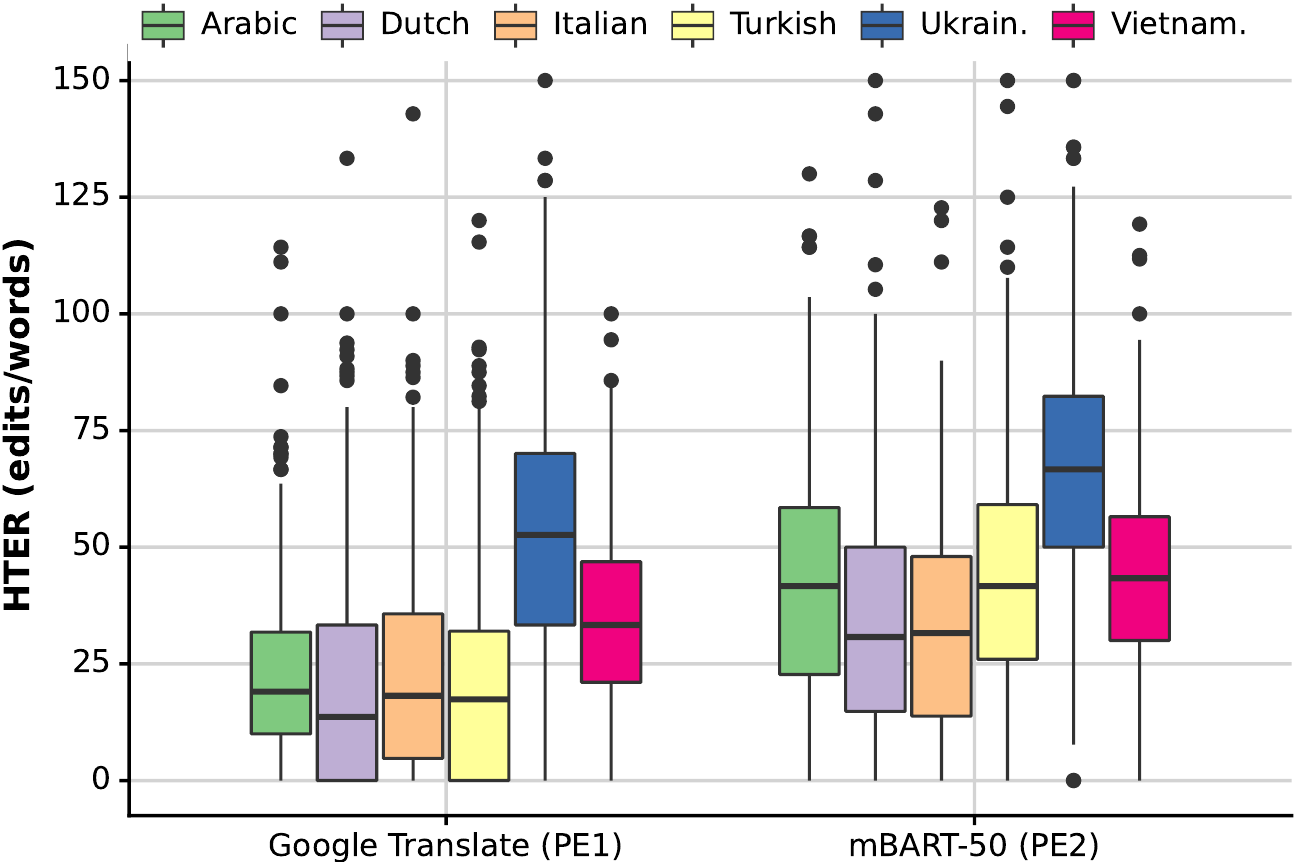}
  \caption{Human-targeted Translation Edit Rate (HTER) for Google Translate and mBART-50 post-editing across available languages.}
  \label{fig:hter}
\end{figure}

\begin{figure}[t]
    \centering
    \includegraphics[width=\linewidth]{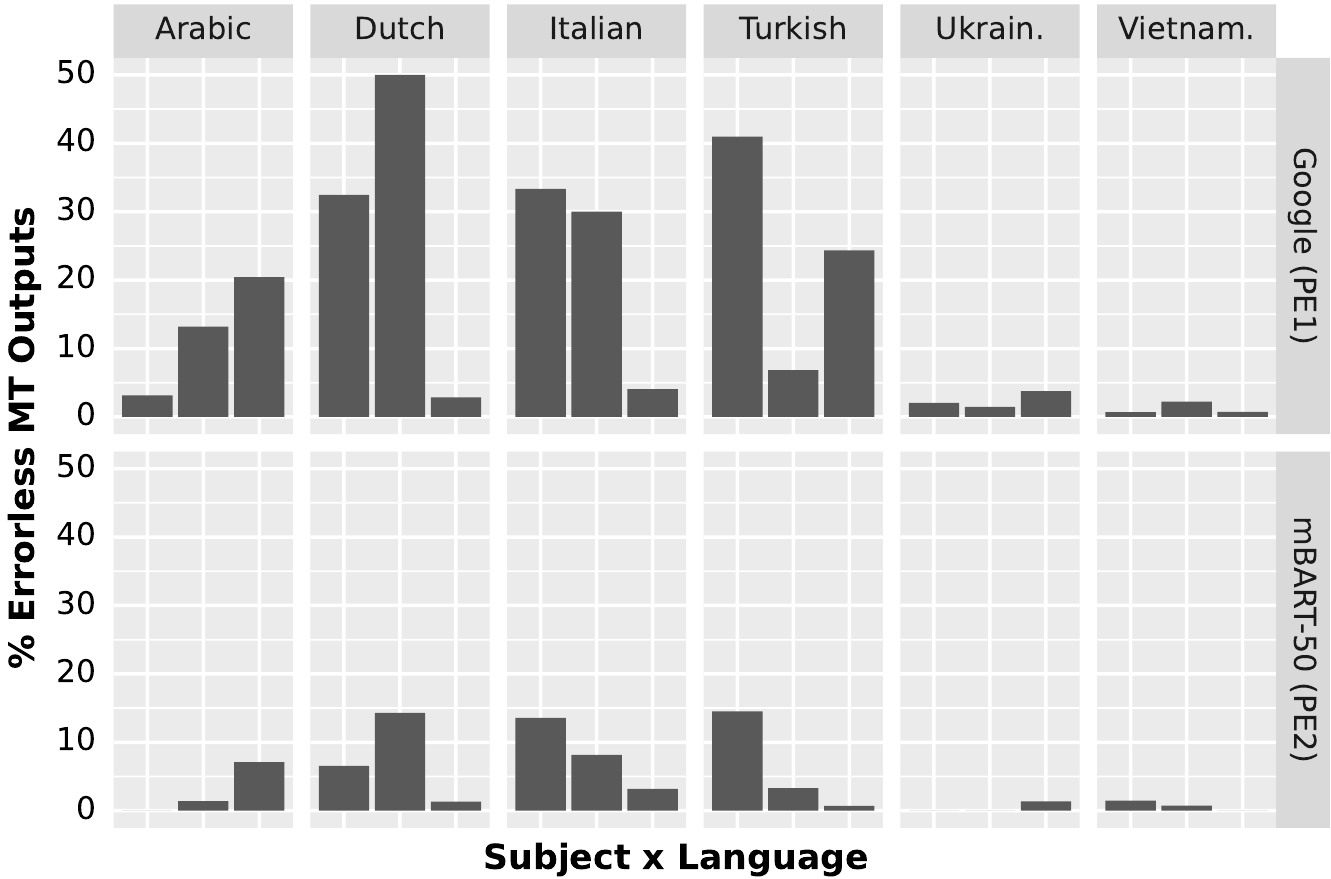}
  \caption{Distribution of error-less machine translation sentence outputs (no edits performed during post-editing) for each translator and every language.}
  \label{fig:errorless-sentences}
\end{figure}

\subsection{Post-Editing Rate}
\label{sec:pe-rate}

We proceed to study the post-editing patterns using the widely-adopted Human-targeted Translation Edit Rate (HTER,~\citeauthor{snover-etal-2006-study}~\citeyear{snover-etal-2006-study}), computed as the length-normalized sum of word-level substitutions, insertions, deletions and shift operations performed during post-editing.\footnote{See Appendix~\ref{app:other} for extra results with a character-level variant of HTER.}

As shown in Figure~\ref{fig:hter}, PE$_1$ required less editing than PE$_2$ for all languages, and a high variability is observed across the two systems and all languages. Since translators were not informed about the presence of two MT systems, we exclude the possibility that these results reflect an over-reliance or distrust towards a specific MT system.
For Google Translate, Ukrainian shows the heaviest edit rate, followed by Vietnamese, whereas Arabic, Dutch, Italian and Turkish all show relatively low amounts of edits. 
Focusing again on mBART-50 for a fairer cross-lingual comparison, Ukrainian is by far the most heavily edited language, followed by a medium-tier group composed of Vietnamese, Arabic and Turkish, and finally by Dutch and Italian as low-edit languages. Results show that several of our observations on the linguistic relatedness and type of morphology also apply to edit rates, with languages less related to English or having richer morphology requiring more post-edits on average.

Figure~\ref{fig:errorless-sentences} visualizes the large gap in edit rates across languages and subjects by presenting the amount of ``errorless'' MT sentences that were accepted directly, i.e. without any post-editing. We note again how the NMT system heavily influences the rate of occurrence of such sentences but nonetheless shows how Dutch and Italian generally present more errorless sentences than Ukrainian and Vietnamese. In particular, for Google Translate outputs, the average rate of error-less sentences is roughly 25\% for the former target languages, while for the latter, it accounts only for the 3\% of total translations. Surprisingly, the English-Turkish pair also fares well, despite the low source-target relatedness.

Finally, we note that post-editing effort appears to correlate poorly with the automatic MT quality metrics reported in Table~\ref{tab:flores-perf} (e.g. see high scores of Vietnamese and low scores of Dutch PE$_1$), highlighting a difficulty in predicting the benefits of MT post-editing over HT for new language pairs.

\begin{figure}[t]
    \centering
    \includegraphics[width=\linewidth]{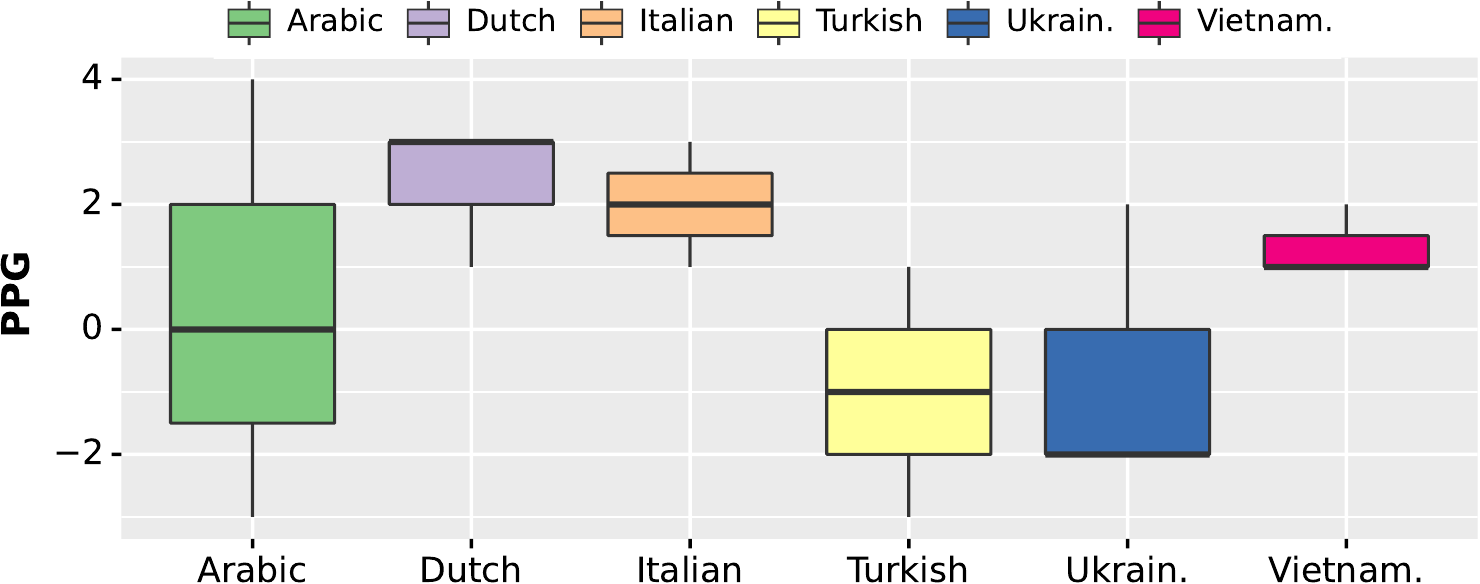}
  \caption{Perceived productivity gains (PPG) between the HT and PE translation modalities, assessed for all subjects after task completion.}
  \label{fig:perceived-pe-speedup}
\end{figure}

\subsection{Perception of Productivity Gain}
We conclude our analysis by examining the post-task questionnaires, in which participants expressed their perception of MT quality and translation speed across HT and PE modalities (HT$_s$, PE$_s$)\footnote{We reemphasize that subjects were unaware of the presence of two distinct MT systems.} using a 1-7 Likert scale (1 slowest, 7 fastest). We use these to compute the Perceived Productivity Gain (PPG) as $\textrm{PPG} = \textrm{PE}_s - \textrm{HT}_s $
and visualize it in Figure~\ref{fig:perceived-pe-speedup}. We observe that Italian and Dutch, the only target languages with marked productivity gains ($\Delta$HT) regardless of the PE system in Table~\ref{tab:productivity}, are also the only ones having consistently high ($\geq 2$) PPG scores across all subjects. 
Moreover, we remark how PPG for target languages with a large gap in $\Delta$HT scores between high-PE$_1$ and low-PE$_2$ (Arabic, Ukrainian) are hardly distinguishable from those of languages in which $\Delta$HT is low for both PE systems (Turkish, Vietnamese). Notably, 4 out of 18 subjects attribute negative PPGs to the PE modality, even though productivity gains were reported across all subjects and languages.
These results suggest that worst-case usage scenarios may play an important role in driving PPG, i.e. that \textit{subjects' perception of quality is largely shaped by particularly challenging or unsatisfying interactions with the NMT system, rather than the average case}.
Finally, from the post-task questionnaire, PPG scores exhibit a strong positive correlation with the perception of MT adequacy ($\rho$=0.66), fluency ($\rho$=0.46) and overall quality ($\rho$=0.69), and more generally with a higher enjoyability of PE ($\rho$=0.60), while being inversely correlated with the perception of problematic mistranslations ($\rho$=$-$0.60).

\section{Conclusions}

In this work we introduced DivEMT, the outcome of a post-editing study spanning two state-of-the-art NMT systems, 18 professional translators and six typologically diverse target languages under a unified setup. 

We leveraged DivEMT's behavioral data to perform a controlled cross-language analysis of NMT post-editing effort along its temporal and editing effort dimensions.
The analysis reveals that NMT drives significant improvements in productivity across all the evaluated languages, but the magnitude of these improvements depends heavily on the language and the underlying NMT system.
In this setting, productivity measurements across modalities were found to be generally consistent with the recorded editing patterns. Our results indicate that translators working on language pairs with significant post-editing productivity gains, on average, perform fewer edits and accept more machine-generated translations without any editing.
We also observed a disconnect between post-editing productivity gains and MT quality metrics collected for the same NMT systems. Finally, low source-language relatedness and target morphological complexity seem to hinder productivity when NMT is adopted, even in settings where system architecture and training data are controlled for.

In our qualitative analysis, translators' perception of post-editing usefulness was found to be strongly shaped by problematic mistranslations. Languages showing large productivity gains for both NMT systems were the only ones associated with a positive perception of PE-mediated gains, as opposed to mixed or negative opinions for other translation directions.

In future work, a more fine-grained analysis of the types of edits conducted by the translators, and their differences across languages, could shed more light on our current findings. 

\section{Limitations}
\label{sec:limitations}
The subjective component introduced by the presence of multiple translators is an important confounding factor in our setup, especially due to the relatively small number of subjects for each language. In our study, we tried to balance a thorough control of other noise components with a faithful reproduction of a realistic translation scenario. However, we realize that the combination of limited document context provided by FLORES-101, the variety of topics covered in the texts and the experimental nature of the \textsc{Pet} platform constitutes an atypical setting that may have impacted the translators' natural productivity. % of the subjects. 
Moreover, variability in the content of mBART-50 fine-tuning data, despite the comparable sizes, may have played a role in the variability observed for automatic MT evaluation and PE gains across languages. 

\section{Broader Impact and Ethical Considerations}

This line of research aims at providing a more precise and faceted understanding of translation and editing effort across multiple languages, and as such is worth pursuing to ensure a fairer compensation to translators if compared to one-size-fits-all approaches based on automatic quality metrics.
Furthermore, the understanding of the application of MT to translators' work in less researched languages and the diversity of measures obtained can give a clearer picture of MT usability, in its broader sense, than automatic metrics. It is relevant to test NMT models in controlled translation environments.
In our experiment, Language Service Providers were paid their requested rate. All words were paid as new words, as the MT usability was unknown prior to the experiment. They were also given thorough instructions and ample time to complete the assignment, accommodating for the COVID-19 pandemic that affected some of the participants. Translators were informed that they could opt-out at any time and have their information deleted.

\section*{Acknowledgements}
Data collection was fully funded by the Dutch Research Council (NWO) under project number 639.021.646. GS is supported by the NWO project InDeep (NWA.1292.19.399). AGA was supported by the European Union’s Horizon 2020 research and innovation program under the Marie Skłodowska-Curie grant agreement No. 890697.
We thank the Center for Information Technology of the University of Groningen for providing access to the Peregrine HPC cluster.

\bibliography{anthology,custom}
\bibliographystyle{emnlp22}

\appendix

\clearpage

\section{Modality scheduling}
\label{app:modality-scheduling}

Table~\ref{tab:schedule} shows an example of the adopted modality scheduling. The modality of document docM$_i$ for translator T$_j$ in the main task is picked randomly among the two modalities that were  not seen by the same translator for docM$_{i-1}$, enforcing consecutive documents given to the same translator to be assigned different modalities to avoid periodicity in repetition and enable the same-language comparisons of Section~\ref{sec:pe_effort}. Importantly, although all three modes were collected for every document, we did not enforce mode consistency across the same translator identifier across languages (i.e. T$_1$ for Italian does not have the same sequence of modalities of translator T$_1$ in Arabic, for example). For this reason, individual subjects are not directly comparable across languages. This is relevant since, e.g. T$_3$ for Dutch and Italian did not operate on the same set of sentences on the same modalities, and thus their comparable editing behavior in Figure~\ref{fig:errorless-sentences} should be attributed to personal preference rather than an identical assignment of modalities on the same sentences. Despite modality scheduling, we have no guarantees that translators consistently follow the order of documents presented in \textsc{Pet}, and thus possibly operate on documents assigned to the same modality consecutively. However, this possibility reduces to random guessing due to a lack of any identifying information related to the modality until the document is entered for editing. The sequence of modalities for the warmup task is fixed and is: HT, PE$_2$, PE$_1$, HT, PE$_2$.

\section{Subject Information}
\label{app:subject-info}

During the setup of our experiment, one translator refused to carry out the main task after the warmup phase, and another was substituted by our choice. Both translators were working in the English-Italian direction and were found to make heavy usage of copy-pasting during the warmup stage, suggesting an incorrect utilization of the platform in light of our guidelines. Both translators, which we identified as T$_2$ and T$_3$ for Italian, were replaced by T$_5$ and T$_4$ respectively. Table~\ref{tab:subjects-info} reflects the final translation selection for all languages, with the information collected by means of the pre-task questionnaire.

\begin{table}[t]\centering\small
  \begin{tabular}{c l l l l}
\toprule
     & & T$_1$ & T$_2$ & T$_3$ \\
\midrule
\parbox[t]{2mm}{\multirow{4}{*}{\rotatebox[origin=c]{90}{warm-up}}} 
 & docW$_1$ & HT & HT & HT \\
 & docW$_2$ & PE$_1$ & PE$_1$ & PE$_1$ \\
 & \ \ ... &  &  & \\
 & docW$_N$ & PE$_2$ & PE$_2$ & PE$_2$ \\
\midrule
\parbox[t]{2mm}{\multirow{4}{*}{\rotatebox[origin=c]{90}{main}}} 
 & docM$_1$ & HT & PE$_1$ & PE$_2$ \\
 & docM$_2$ & PE$_2$ & HT & PE$_1$ \\
 & docM$_3$ & HT & PE$_2$ & PE$_!$ \\
 & \ \ ... &  &  & \\
 & docM$_N$ & PE$_2$ & PE$_1$ & HT \\
\bottomrule
\end{tabular}
  \caption{Modality scheduling overview. For each language, each subject (T$_i$) works with a pseudo-random sequence of modalities (HT, PE$_1$, PE$_2$). For the warm-up task ($N=5$), all translators are provided with the same documents in the same modalities. For the main task ($N=107$), each translator is assigned a modality at random. Each document is translated once for every modality. The same procedure is repeated independently for all the languages.}
  \label{tab:schedule}
\end{table}

\begin{table*}[!ht]
    \small
    \centering
    \begin{tabular}{lccccccccc}
        \toprule
         & & \textbf{Gender} & \textbf{Age} & \textbf{Degree} & \textbf{Position} & \textbf{En Level} & \textbf{YoE} & \textbf{YoE w/ PE} & \textbf{\% PE}  \\
        \midrule
        \multirow{3}{*}{\textbf{Arabic}}
          & T$_1$ & M & 35-44 & BA & Freelancer & C2 & > 15  & 2-5  & 20\%-40\% \\ 
          & T$_2$ & M & 25-34 & BA & Employed   & C2 & 5-10  & 2-5  & 60\%-80\% \\ 
          & T$_3$ & M & 25-34 & MA & Freelancer & C1 & 5-10  & < 2  & 20\%-40\% \\
        \midrule
        \multirow{3}{*}{\textbf{Dutch}} 
          & T$_1$ & M & 25-34 & MA & Freelancer & C2 & 5-10  & 5-10  & 60\%-80\% \\ 
          & T$_2$ & F & 35-44 & MA & Freelancer & C1 & 10-15 & 5-10  & 40\%-60\% \\ 
          & T$_3$ & F & 25-34 & MA & Freelancer & C2 & 2-5   & 2-5   & 20\%-40\% \\
        \midrule
        \multirow{3}{*}{\textbf{Italian}} 
          & T$_1$ & F & 25-34 & MA & Employed   & C1 & 5-10   & 5-10 & 20\%-40\% \\ 
          & T$_5$ & F & 25-34 & MA & Freelancer & C1 & 2-5    & 2-5  & 40\%-60\% \\ 
          & T$_4$ & F & 35-44 & BA & Freelancer & C2 & 10-15  & 5-10 & > 80\%    \\
        \midrule
        \multirow{3}{*}{\textbf{Turkish}} 
          & T$_1$ & F & 25-34 & BA          & Freelancer & C2 & 5-10   & 2-5  & < 20\% \\ 
          & T$_2$ & F & 25-34 & BA          & Freelancer & C1 & 5-10   & 5-10 & < 20\% \\ 
          & T$_3$ & M & 25-34 & High school & Freelancer & C2 & 10-15  & < 2  & < 20\% \\
        \midrule
        \multirow{3}{*}{\textbf{Ukrainian}} 
          & T$_1$ & F & 35-44 & MA          & Employed & C1 & 5-10  & 5-10  & 20\%-40\% \\ 
          & T$_2$ & M & 35-44 & MA          & Employed & C1 & 10-15 & 10-15 & 20\%-40\% \\ 
          & T$_3$ & M & 35-44 & High school & Employed & B2 &  2-5  &  2-5  & 20\%-40\% \\
        \midrule
        \multirow{3}{*}{\textbf{Vietnamese}} 
          & T$_1$ & F & 25-34 & MA & Employed   & C2 & 10-15 & 5-10  & 40\%-60\% \\ 
          & T$_2$ & F & 25-34 & BA & Freelancer & C1 & 5-10  & < 2   & 20\%-40\% \\ 
          & T$_3$ & F & 25-34 & MA & Employed   & C1 & 2-5   & < 2   & < 20\% \\
        \bottomrule
    \end{tabular}
  \caption{Subjects information for DivEMT. The last three columns represent respectively the number of years of professional experience as a translator (YoE), the number of years of experience with MT post-editing (YoE w/ PE) and the \% of work assignments requiring post-editing in the last 12 months (\% PE) for each subject.}
  \label{tab:subjects-info}
\end{table*}

\begin{table}\centering\small
  \begin{tabular}{l c c c c c}
\toprule
 Type & WN & WV & WB & \# Sent. & \# Words \\
\midrule
 3S   & 11 & 13 & 11 & 105 & 2168 \\
 4S   & 14 & 8  & 13 & 140 & 3214 \\
 5S   & 12 & 13 & 12 & 185 & 3826 \\
 Tot. & 37 & 34 & 36 & 450 & 9626 \\
\bottomrule
\end{tabular}
  \caption{Distribution of the selected DivEMT documents across sizes and Wikipedia categories. A Type value of $N$S stands for documents composed by $N$ contiguous sentences, WN, WV and WB stand respectively for WikiNews, WikiVoyage and Wikibooks}
  \label{tab:sources}
\end{table}

\section{Translation Guidelines}\label{app:guidelines}

An extract of the translation guidelines provided to the translators follows. The full guidelines are provided in the additional materials.

\small\vspace{3.5mm}

Fill in the pre-task questionnaire before starting the project. In this experiment, your goal is to complete the translation of multiple files in one of two possible translation settings. Please, complete the tasks on your own, even if you know another translator that might be working on this project. The translation setting alternates between texts, with each text requiring a single translation in the assigned setting. The two translation settings are:
\begin{enumerate}
    \item Translation from scratch. Only the source sentence is provided, you are to write the
translation from scratch.
    \item Post-editing. The source sentence is provided alongside a translation produced by
an MT system. You are to post-edit this MT output. Post-edit the text so you are
satisfied with the final translation (the required quality is publishable quality). If
the MT output is too time-consuming to fix, you can delete it and start from scratch.
However, please do not systematically delete the provided MT output to give your
own translation.
\end{enumerate}
    
Important: All editing MUST happen in the provided PET interface: that is, working in other editors and copy-pasting the text back to PET is NOT ALLOWED, because it invalidates the experiment. This is easy to spot in the log data, so please avoid doing this. Complete the translation of all files sequentially, i.e. in the order presented in the tool. DO NOT SKIP files at your own convenience. Make sure that ALL files are translated when you deliver the tasks.

The aim is to produce publishable professional quality translations for both translation settings. Thus, please translate to your best abilities. You can return to the files and self-review as many times as you think it is necessary. Important: The time invested to translate is recorded while the active unit (sentence) is in editing mode (yellow background). Therefore:
    \begin{itemize}
        \item Only start to translate when you are in editing mode (yellow background). In other
words, do not start thinking how you will translate a sentence when the active unit
is not yet in editing mode (green or red background).
        \item Do not leave a unit in editing mode (yellow background) while you do something
else. If you need to do something unrelated in the middle of a translation then go
out of editing mode and come back to editing mode when you are ready to resume
translating.
    \item First you will be translating a warmup task, and then the main task. When you are
translating each file, you can consult the Source text (ST) by looking up the url in the
Excel files that we have sent for reference.
    \end{itemize}
In order to find the correct terminology for the translation you can consult any source in the Internet. Important: However, it is NOT ALLOWED to use any MT engine to find terms or alternatives to translations (such as Google Translate, DeepL, MS Translator or any MT engine available in your language). Using MT engines invalidates the experiment, and will be detected in the log data. Please fill-in the post-task questionnaire ONLY ONCE after completing all the translation tasks (both warmup and main tasks).

\normalsize

\begin{table*}[ht]
    \centering\footnotesize
    \begin{tabular}{p{1.2in}p{\textwidth - 1.8in}}
    \toprule
        \textbf{Field name} & \textbf{Description} \\
        \midrule
        \scriptsize\texttt{unit\_id, flores\_id, subject\_id, task\_type} & Identifiers for the item, respective FLORES-101 sentence, translator and translation mode. \\
        \midrule
        \scriptsize\texttt{src\_text} & The original source sentence extracted from Wikinews, wikibooks or wikivoyage. \\
        \scriptsize\texttt{mt\_text} & MT output sentence before post-editing, present only if \texttt{task\_type} is `pe'. \\
        \scriptsize\texttt{tgt\_text} & Final sentence produced by the translator (either from scratch or post-editing \texttt{mt\_text}) \\
        \scriptsize\texttt{aligned\_edit} & Aligned visual representation of the machine translation and its post-edit with edit operations \\
        \midrule
        \scriptsize\texttt{edit\_time} & Total editing time for the translation in seconds. \\
        \midrule
        \scriptsize\texttt{k\_letter, k\_digit, k\_white, k\_symbol, k\_nav} & Number of keystrokes for various key types (letters, digits, keystrokes, whitespaces, punctuation, navigation keys) during the translation. \\
        \scriptsize\texttt{k\_erease, k\_copy, k\_paste, k\_cut, k\_do} & Number of keystrokes for erease (backspace, cancel), copy, paste, cut and Enter actions during the translation. \\
        \scriptsize\texttt{k\_total} & Total number of all keystroke categories during the translation. \\
        \midrule
        \scriptsize\texttt{n\_pause\_geq\_N}, \texttt{len\_pause\_geq\_N} & Number and length of pauses longer than 300ms and 1000ms during the translation. \\
        \midrule
        \scriptsize\texttt{num\_annotations} & Number of times the translator focused the target sentence texbox during the session. \\
        \midrule
        \scriptsize\texttt{n\_insert, n\_delete, n\_substitute, n\_shift, tot\_shifted\_words, tot\_edits, hter} & Granular editing metrics and overall HTER computed using the Tercom library. \\
        \scriptsize\texttt{cer} & Character-level HTER score computed between the MT and post-edited outputs. \\
        \midrule
        \scriptsize\texttt{bleu, chrf} & Sentence-level BLEU and ChrF scores between MT and post-edited fields computed using the SacreBLEU library with default parameters. \\
        \midrule
        \scriptsize\texttt{time\_per\_char, key\_per\_char, words\_per\_hour, words\_per\_minute} & Edit time per source character, expressed in seconds. Proportion of keys per character needed to perform the translation. Amount of source words translated or post-edited per hour/minute \\
        \midrule
        \scriptsize\texttt{subject\_visit\_order} & Id denoting the order in which the translator accessed documents in the interface. \\
        \bottomrule
    \end{tabular}
  \caption{Description of the main fields associated to every DivEMT data entry. An entry correspond to a translation in a specific modality (HT, PE$_1$ or PE$_2$) for one of the six target languages}
  \label{tab:divemt-fields}
\end{table*}

\section{Details on Document Selection and Preprocessing}
\label{app:doc-select}

\begin{figure}
    \centering
    \includegraphics[width=0.8\linewidth]{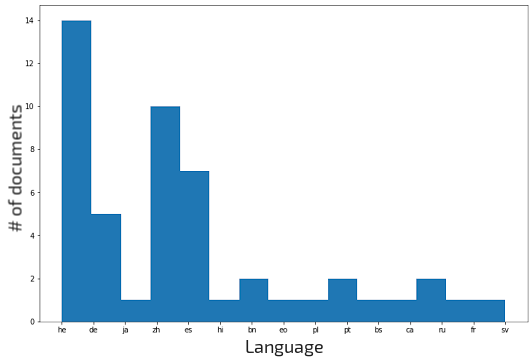}
    \includegraphics[width=0.8\linewidth]{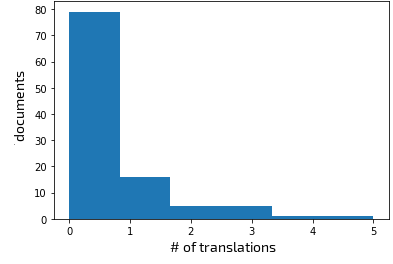}
  \caption{Top: Distribution for the availability of documents selected for DivEMT in languages other than English. Bottom: Quantity of selected documents per number of available translations of Wikipedia.}
  \label{fig:lang-doc-select}
\end{figure}

\paragraph{Document selection} Table~\ref{tab:sources} present the distribution of selected documents from the Flores-101 devtest split based on their domain and the number of sentences that compose them. The first goal in the selection process was to preserve a rough balance between the three categories while including mostly 4 and 5-sentence docs which are faster to edit in \textsc{Pet} (no need to frequently close and reopen an editing window). Another objective of the selection was to minimize the chance of translators finding the translated version of the Wikipedia article from which documents were taken and copied from there, despite our guidelines. We thus scrape the articles from Wikipedia and assess the number of available translations. Among the selected documents, only a small subset has translations in other languages (see Figure~\ref{fig:lang-doc-select} top, an article can have multiple languages), mainly in Hebrew (14), Chinese (10), Spanish (7) and German (5) respectively. Considering the total number of translations for every article (Figure~\ref{fig:lang-doc-select} bottom), we see that roughly 75\% of them (79 docs) have no translations. We consider this satisfactory as proof there should not be a large amount of possible copying involved, and we follow up on this evaluation by also ensuring that no repeated copy-paste patterns are present in keylogs after the warmup stage.

\paragraph{Filtering of Outliers} For our analysis of Section~\ref{sec:pe_effort}, we only use sentences with an editing time lower than 45 minutes, which was selected heuristically as a reasonably high threshold to allow for extensive searching and thinking. In the following, we present the identifiers of the sentences that were filtered out during this process. E.g. 54.1 means the first sentence of document 54, having \texttt{item\_id} equal to \texttt{flores101-main-541} in the dataset. Note that the sentences were outliers only for 2/6 languages and were all different, indicating no systematic issues in the sample: \textsc{Ara}: 54.1, 100.3, \textsc{Vie}: 3.1, 3.2, 24.3, 28.4, 33.1, 33.2, 40.3, 41.2, 50.3, 100.1, 102.1, 106.1, 107.2, 107.4. The 17 sentences were removed for all modalities and languages in the analysis of Section~\ref{sec:pe_effort} to preserve the validity of our comparison, representing a loss of roughly 4\% of the total available data, a tolerable amount for our analysis.

\paragraph{Fields Description} Table \ref{tab:divemt-fields} presents the set of fields that were collected for every entry of the DivEMT dataset. The fields related to keystrokes, times, pauses, annotations and visit order were extracted from the event log of \textsc{Pet} .per files, while edits information and other MT quality metrics were computed in a second moment with the help of widely-used libraries.

\paragraph{Additional Notes on \textsc{Pet}} The \textsc{Pet} platform was modified to enable a correct right-to-left language visualization, which was necessary for Arabic.

\section{Other Measurements}
\label{app:other}

\begin{figure}[t]
    \centering
    \includegraphics[width=\linewidth]{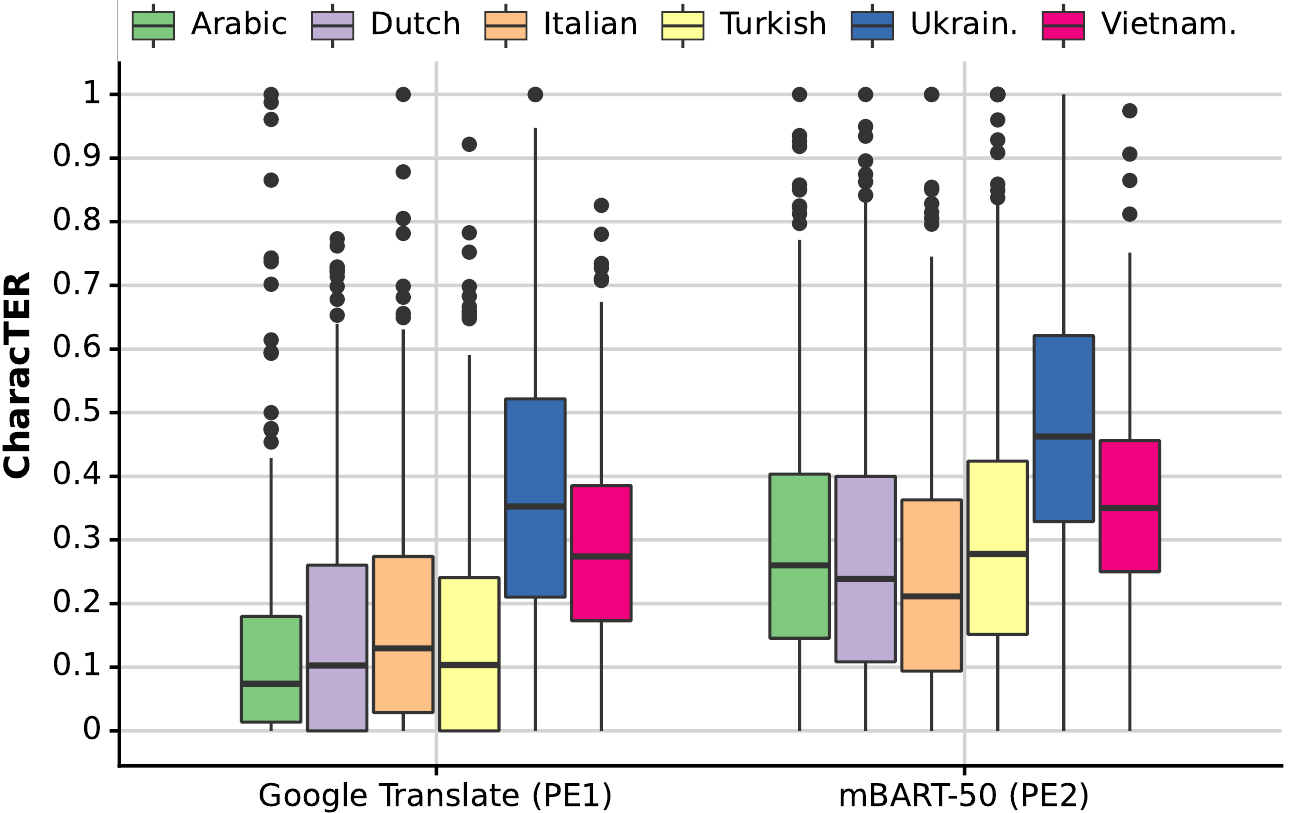}
  \caption{Character-level Human-targeted Translation Edit Rate (CharacTER) for Google Translate and mBART-50 post-editing across available languages.}
  \label{fig:character}
\end{figure}

\paragraph{CharacTER Across Systems and Languages}
While HTER is a standard metric adopted both in academic and industrial settings, we also evaluated its character-level variant CharacTER~\citep{wang-etal-2016-character} to assess whether it could better account for the editing process of morphologically rich languages.
Figure~\ref{fig:character} presents the CharacTER results. When comparing this plot to the HTER one (Figure~\ref{fig:hter}),
we notice that CharacTER preserves the overall trends, but slightly improves the edit rate for Arabic and Turkish with respect to other languages. Nevertheless, we find HTER to correlate slightly better with productivity scores across all tested languages, both at a sentence and at a document level. For this reason, word-level results are reported in the article's main body.

\paragraph{Automatic Evaluation of NMT Systems} The selection of systems used in this study was driven by a broader evaluation procedure covering more models, metrics and target languages. Table~\ref{tab:flores-perf-full} presents the overall results of our evaluation. We use HuggingFace’s Transformers library~\citep{wolf-etal-2020-transformers} for all neural models, using the default decoding settings without further fine-tuning.  All metrics were computed using the default settings of SacreBLEU~\citep{post-2018-call} and Comet~\citep{rei-etal-2020-comet}.

\begin{figure*}
    \centering
    \includegraphics[width=\linewidth]{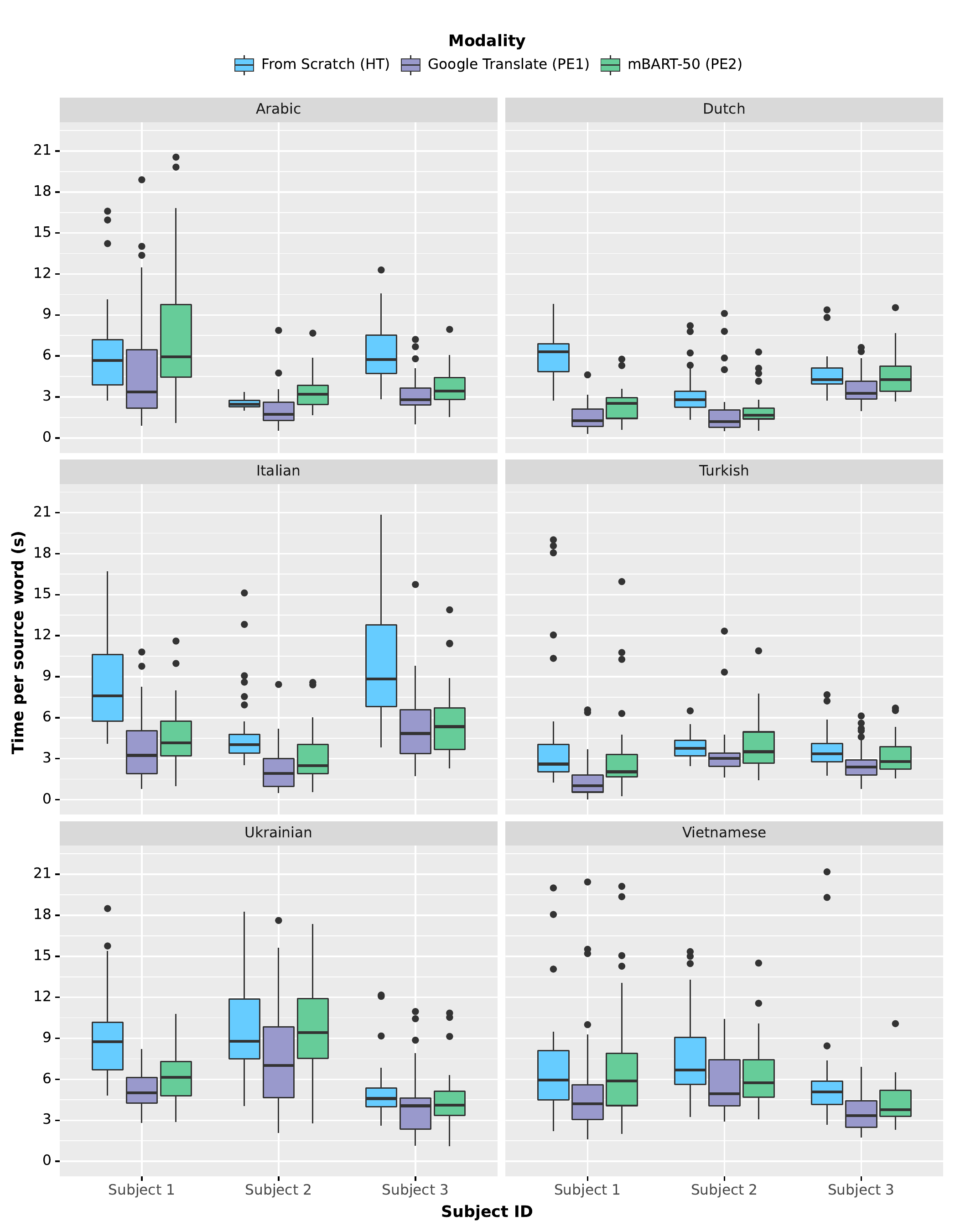}
  \caption{Time per processed source word across languages, subjects and translation modalities, measured in seconds. Each point represents a document containing 3--5 sentences translated by a subject in one of the languages, with higher scores representing slower editing.}
  \label{fig:time-per-src-word-per-translator}
\end{figure*}

\paragraph{Inter-subject Variability in Translation Times} Although the variability across different subjects working on the same language directions is not the main concern of our investigation, we produce Figure~\ref{fig:time-per-src-word-per-translator} (an expanded version of Figure~\ref{fig:time-per-src-word}) to visualize the inter-subject variability for translation times. We observe that the variability across different translators is more pronounced when translating from scratch and that the overall trend of speed improvements associated with PE is mostly preserved (with few exceptions related to the PE$_2$ modality).

\begin{table*}[!ht]
    \small
    \centering
    \begin{tabular}{lcccccc}
        \toprule
         & \textbf{System} & \textbf{BLEU} & \textbf{chrF2} & \textbf{TER} & \textbf{chrF2++} & \textbf{COMET} \\
        \midrule
        \multirow{4}{*}{\textbf{Arabic}}
          & M2M100 & 19.2 & 50.9 & 69.2 & 47 & 0.417 \\ 
          & MarianNMT & \underline{22.7} & \underline{54.2} & \underline{64.7} & \underline{50.4} & \underline{0.483} \\ 
          & mBART-50 & 17 & 48.5 & 69.1 & 44.8 & 0.452 \\
          & GTrans & \textbf{34.1} & \textbf{65.6} & \textbf{52.8} & \textbf{61.9} & \textbf{0.737} \\
        \midrule
        \multirow{5}{*}{\textbf{Dutch}} 
          & M2M100 & 21.3 & 52.9 & 66.1 & 49.8 & 0.405 \\ 
          & MarianNMT & \underline{25} & \underline{56.9} & \underline{62.5} & \underline{53.8} & \underline{0.543} \\ 
          & mBART-50 & 22.6 & 53.9 & 63.7 & 50.9 & 0.532 \\ 
          & DeepL & 28.7 & 59.5 & 59.5 & 56.6 & \textbf{0.67} \\ 
          & GTrans & \textbf{29.1} & \textbf{60} & \textbf{58.5} & \textbf{57.1} & 0.667 \\ 
        \midrule
        \multirow{4}{*}{\textbf{Indonesian}} 
          & M2M100 & 35.9 & 63.1 & 47.3 & 60.8 & 0.614 \\ 
          & MarianNMT & \underline{38.5} & \underline{65.6} & \underline{46.5} & \underline{63.3} & 0.671 \\ 
          & mBART-50 & 35.9 & 63.3 & 47.7 & 61.1 & \underline{0.706} \\ 
          & GTrans & \textbf{51.5} & \textbf{73.6} & \textbf{34.5} & \textbf{71.9} & \textbf{0.894} \\
        \midrule
        \multirow{5}{*}{\textbf{Italian}} 
          & M2M100 & 23.6 & 53.9 & 63.2 & 51 & 0.51 \\ 
          & MarianNMT & \underline{27.5} & \underline{57.6} & \underline{58.9} & \underline{54.8} & 0.642 \\ 
          & mBART-50 & 24.4 & 54.7 & 61.2 & 51.8 & \underline{0.648} \\ 
          & DeepL & \textbf{33} & 61 & 54 & 58.5 & \textbf{0.795} \\ 
          & GTrans & 32.8 & \textbf{61.4} & \textbf{53.6} & \textbf{58.8} & 0.781 \\
        \midrule
        \multirow{4}{*}{\textbf{Japanese}} 
          & M2M100 & 24.5 & 32.2 & 123.3 & 26 & 0.389 \\
          %& MarianNMT & 18.1 & 26.8 & 1548.1 & 20.5 & -  \\ 
          & mBART & \underline{27.1} & \underline{35.4} & \underline{123} & \underline{28.3} & \underline{0.538} \\ 
          & DeepL & \textbf{41.3} & \textbf{46.8} & 108 & \textbf{37} & \textbf{0.75} \\ 
          & GTrans & 38.4 & 44.7 & \textbf{101.5} & 33.9 & 0.683 \\
        \midrule
        \multirow{5}{*}{\textbf{Polish}} 
          & M2M100 & 16.1 & 46.5 & 74.2 & 43.1 & 0.486 \\ 
          & MarianNMT & \underline{19.3} & \underline{49.9} & \underline{70.5} & \underline{46.6} & \underline{0.648} \\ 
          & mBART-50 & 17.4 & 48.2 & 72.4 & 44.9 & 0.603 \\ 
          & DeepL & 24 & 54.3 & 66.4 & 51.1 & \textbf{0.832} \\ 
          & GTrans & \textbf{24.4} & \textbf{54.6} & \textbf{64.6} & \textbf{51.4} & 0.804 \\
        \midrule
        \multirow{5}{*}{\textbf{Russian}} 
          & M2M100 & 22.5 & 51.1 & 65.6 & 48.1 & 0.427 \\ 
          & MarianNMT & \underline{25.4} & \underline{53.5} & 64.3 & \underline{50.7} & 0.537 \\ 
          & mBART & 24.8 & 52.6 & \underline{63.7} & 49.7 & \underline{0.541} \\ 
          & DeepL & \textbf{35.9} & \textbf{61.8} & \textbf{53.3} & \textbf{59.3} & \textbf{0.79} \\ 
          & GTrans & 33 & 60.5 & 55.2 & 57.7 & 0.731 \\
        \midrule
        \multirow{4}{*}{\textbf{Turkish}} 
          & M2M100 & 20.3 & 53.9 & 65.2 & 50.1 & 0.686 \\ 
          & MarianNMT & \underline{26.3} & \underline{59.8} & \underline{58.8} & \underline{55.8} & \underline{0.881} \\ 
          & mBART-50 & 18.8 & 52.7 & 67.5 & 48.7 & 0.755 \\ 
          & GTrans & \textbf{35} & \textbf{65.5} & \textbf{50.4} & \textbf{62.2} & \textbf{1} \\
        \midrule
        \multirow{4}{*}{\textbf{Ukrainian}} 
          & M2M100 & 21.9 & 51.4 & 65.8 & 48.3 & 0.463 \\ 
          & MarianNMT & 20 & 48.8 & 69.2 & 45.7 & 0.427 \\ 
          & mBART-50 & \underline{21.9} & \underline{50.7} & \underline{67.9} & \underline{47.7} & \underline{0.587} \\ 
          & GTrans & \textbf{31.1} & \textbf{59.8} & \textbf{55.9} & \textbf{56.8} & \textbf{0.758} \\ 
        \midrule
        \multirow{4}{*}{\textbf{Vietnamese}} 
          & M2M100 & 33.3 & 52.3 & 52.4 & 52.1 & 0.43 \\ 
          & MarianNMT & 26.7 & 45.7 & 60.2 & 45.6 & 0.117 \\ 
          & mBART-50 & \underline{34.7} & \underline{54} & \underline{50.7} & \underline{53.8} & \underline{0.608} \\ 
          & GTrans & \textbf{45.1} & \textbf{61.9} & \textbf{41.8} & \textbf{61.9} & \textbf{0.724} \\
        \bottomrule
    \end{tabular}
  \caption{Automatic MT quality of all evaluated NMT systems on all tested languages in the English-to-XX setting, using the FLORES-101 full devtest for evaluation. Besides mBART-50 and Google Translate (GTrans), we also evaluate a set of bilingual Transformer-based NMT models trained with MarianNMT~\citep{tiedemann-thottingal-2020-opus}, the DeepL industrial MT system and the multilingual M2M-100 418M model~\citep{fan-etal-2021-beyond}. Overall best performance per language is highlighted in \textbf{bold}, best open-source system performance per language is \underline{underlined}.}
  \label{tab:flores-perf-full}
\end{table*}

%\begin{table*}[ht]\centering
%  \begin{tabular}{@{\ }l@{\ } cc c@{\ }c c@{\ }c c@{\ }}
%\toprule
%& & & \multicolumn{2}{c}{\textbf{Morphology}} & \multicolumn{2}{c}{\textbf{Word Order}} &  \\
%\cmidrule(lr){4-5}
%\cmidrule(lr){6-7}
%& \textbf{Genus} & \textbf{Family} & \textbf{Type} \ \ \ & \textbf{\# Cases} & \textbf{Dominant} & \textbf{Freedom} & %\textbf{Script} \\
%\midrule
%\textbf{English} & Indo-European & Germanic & Fusional & (3) & \textsc{svo} & low & Latin \\  
%\midrule 
%\textbf{Arabic (\textsc{msa})} & Afro-Asiatic & Semitic & Introflex. & 3 & \textsc{vso/svo} & low & Arabic \\  
%\textbf{Dutch} & Indo-European & Germanic & Fusional & -- & \textsc{svo/sov/v2} & low & Latin \\  
%\textbf{Italian} & Indo-European & Romance & Fusional & -- & \textsc{svo} & low & Latin \\  
%\textbf{Turkish} & Altaic & Turkic & Agglutin. & 6 & \textsc{sov} & high & Latin* \\  
%\textbf{Ukrainian} & Indo-European & Slavic & Fusional & 7 & \textsc{svo} & high & Cyrillic \\  
%\textbf{Vietnamese} & Austro-Asiatic & Viet-Muong & Isolating & -- & \textsc{svo} & low & Latin* \\  
%\bottomrule
%\end{tabular}
%  \caption{The typological diversity of the selected language sample across various linguistic dimensions. Note that %English has 3  cases, but these are only marked for pronouns. V2 indicates verb-second word order. Latin* indicates %language-specific extensions of the Latin alphabet. Sources include the World Atlas of Language Structures \cite{wals} %and \cite{futrell-15-quantifying}.}
%  \label{tab:languages}
%\end{table*}

\section{Full DivEMT Examples}
\label{app:examples}

Tables~\ref{tab:data_example_full_1} and~\ref{tab:data_example_full_2} present two full examples of DivEMT entries, including all output modalities, intermediate MT outputs, post-edits and edit highlights for all target languages.

\begin{table*}
    \centering
    \footnotesize
    \begin{tabular}{p{0.2in} p{\textwidth - 0.6in}}
        \toprule
        \textsc{English} \\
        & \vspace{0.3mm} Inland waterways can be a good theme to base a holiday around. \\
        \midrule
        \textsc{Arabic} \\ \vspace{0.2mm}
        HT & \vspace{-1.5mm} \setcode{utf8}{\< يمكن أن تكون الممرات المائية الداخلية خياراً جيدًا لتخطيط عطلة حولها. >}\\
        \cmidrule(lr){2-2}
        \multirow{2}{*}{PE$_1$} 
        & \textbf{MT:} \includegraphics[height=5mm]{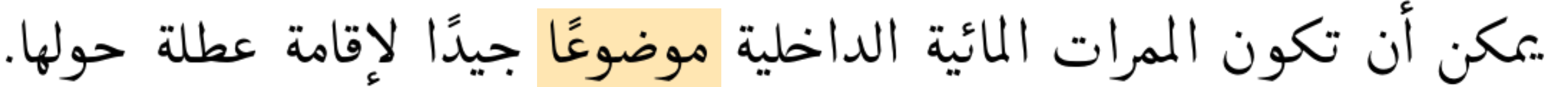} \\%\< يمكن أن تكون الممرات المائية الداخلية موضوعًا جيدًا لإقامة عطلة حولها.> \\
        & \textbf{PE:} \includegraphics[height=5mm]{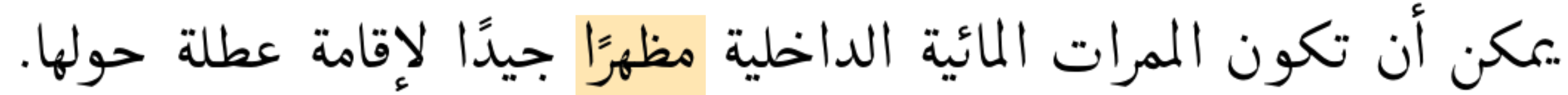} \\%\< يمكن أن تكون الممرات المائية الداخلية مظهرًا جيدًا لإقامة عطلة حولها. >\\
        \cmidrule(lr){2-2}
        \multirow{2}{*}{PE$_2$}
        & \textbf{MT:} \includegraphics[height=5mm]{MTPE2ARA.pdf} \\%\< يمكن أن تكون السكك الحديدية الداخلية موضوعًا جيدًا لإقامة عطلة حول. >\\
        & \textbf{PE:} \includegraphics[height=5mm]{PEPE2ARA.pdf} \\%\< قد تكون الممرات المائية الداخلية مكانًا جيدًا لقضاء عطلة حولها.>\\
        \midrule
        \textsc{Dutch} \\ \vspace{0.3mm}
        HT & \vspace{0.3mm} Binnenlandse waterwegen kunnen een goed thema zijn voor een vakantie. \\
        \cmidrule(lr){2-2}
        \multirow{2}{*}{PE$_1$} & \textbf{MT:} \colorbox{lightred}{De} binnenwateren kunnen een \colorbox{lightred}{goed thema zijn om een vakantie} \colorbox{lightorange}{omheen te baseren}. \\
        & \textbf{PE:} Binnenwateren kunnen een \colorbox{lightorange}{goede vakantiebestemming zijn}. \\
        \cmidrule(lr){2-2}
        \multirow{2}{*}{PE$_2$} & \textbf{MT:} Binnenwaterwegen kunnen een goed thema zijn om een vakantie rond te \colorbox{lightorange}{zetten}. \\
        & \textbf{PE:} Binnenwaterwegen kunnen een goed thema zijn om een vakantie rond te \colorbox{lightorange}{organiseren}. \\
        \midrule
        \textsc{Italian} \\
        \vspace{0.3mm}
        HT & \vspace{0.3mm} I corsi d'acqua dell'entroterra possono essere un ottimo punto di partenza da cui organizzare una vacanza. \\
        \cmidrule(lr){2-2}
        \multirow{2}{*}{PE$_1$} & \textbf{MT:} \colorbox{lightorange}{Trasporto fluviale può} essere un \colorbox{lightorange}{buon tema} per \colorbox{lightred}{basare} \colorbox{lightblue}{una} \colorbox{lightorange}{vacanza in giro}. \\
        & \textbf{PE:} \colorbox{lightgreen}{I canali di} \colorbox{lightorange}{navigazione interna possono} essere un \colorbox{lightorange}{ottimo motivo} per \colorbox{lightorange}{cui intraprendere} \colorbox{lightblue}{una} \colorbox{lightorange}{vacanza}. \\
        \cmidrule(lr){2-2}
        \multirow{2}{*}{PE$_2$} & \textbf{MT:} I corsi d'acqua interni possono essere un buon tema \colorbox{lightorange}{per fondare} una vacanza. \\
        & \textbf{PE:} I corsi d'acqua interni possono essere un buon tema \colorbox{lightgreen}{su} \colorbox{lightorange}{cui basare} una vacanza. \\
        \midrule
        \textsc{Turkish} \\ \vspace{0.3mm}
        HT & \vspace{0.3mm} İç bölgelerdeki su yolları, tatil planı için iyi bir tema olabilir. \\
        \cmidrule(lr){2-2}
        \multirow{2}{*}{PE$_1$} 
        & \textbf{MT:} İç su yolları, bir \colorbox{lightorange}{tatili temel almak} için iyi bir tema olabilir.\\
        & \textbf{PE:} İç su yolları, bir \colorbox{lightorange}{tatil planı yapmak} için iyi bir tema olabilir.\\
        \cmidrule(lr){2-2}
        \multirow{2}{*}{PE$_2$}
        & \textbf{MT:} İç \colorbox{lightorange}{suyolları,} tatil için uygun bir tema olabilir.\\
        & \textbf{PE:} İç \colorbox{lightorange}{sular} tatil için uygun bir tema olabilir.\\
        \midrule
        \textsc{Ukrainian} \\ \vspace{0.3mm}
        HT & \vspace{0.3mm} \foreignlanguage{ukrainian}{Можна спланувати вихідні, взявши за основу подорож внутрішніми водними шляхами.} \\
        \cmidrule(lr){2-2}
        \multirow{2}{*}{PE$_1$} 
        & \textbf{MT:} \foreignlanguage{ukrainian}{Внутрішні водні шляхи можуть стати гарною темою для \colorbox{lightorange}{відпочинку навколо}.} \\
        & \textbf{PE:} \foreignlanguage{ukrainian}{Внутрішні водні шляхи можуть стати гарною темою для \colorbox{lightorange}{проведення вихідних}.} \\
        \cmidrule(lr){2-2}
        \multirow{2}{*}{PE$_2$}
        & \textbf{MT:} \foreignlanguage{ukrainian}{\colorbox{lightorange}{Водні шляхи можуть} \colorbox{lightblue}{бути} \colorbox{lightorange}{хорошим об 'єктом} \colorbox{lightblue}{для} \colorbox{lightorange}{базування відпочинку навколо}.} \\
        & \textbf{PE:} \foreignlanguage{ukrainian}{\colorbox{lightgreen}{Місцевість} \colorbox{lightorange}{навколо внутрішніх водних шляхів може} \colorbox{lightblue}{бути} \colorbox{lightorange}{гарним вибором} \colorbox{lightblue}{для} \colorbox{lightorange}{організації відпочинку.}} \\
        \midrule
        \textsc{Vietnamese} \\ \vspace{0.3mm}
        HT & \vspace{0.3mm}
        \foreignlanguage{vietnamese}{Du lịch trên sông có thể là một lựa chọn phù hợp cho kỳ nghỉ.} \\
        \cmidrule(lr){2-2}
        \multirow{2}{*}{PE$_1$} 
        & \textbf{MT:} \foreignlanguage{vietnamese}{Đường thủy nội địa có thể là một \colorbox{lightorange}{chủ đề} hay để \colorbox{lightorange}{tạo cơ sở} cho \colorbox{lightred}{một} kỳ \colorbox{lightred}{nghỉ xung} \colorbox{lightorange}{quanh}.} \\
        & \textbf{PE:} \foreignlanguage{vietnamese}{Đường thủy nội địa có thể là một \colorbox{lightorange}{ý tưởng} hay để \colorbox{lightorange}{lập kế hoạch} cho kỳ nghỉ.} \\
        \cmidrule(lr){2-2}
        \multirow{2}{*}{PE$_2$}
        & \textbf{MT:} \foreignlanguage{vietnamese}{\colorbox{lightorange}{Các tuyến nước} \colorbox{lightblue}{nội địa} \colorbox{lightorange}{có thể} là một \colorbox{lightred}{chủ đề tốt} \colorbox{lightorange}{để xây dựng một kì nghỉ.}} \\
        & \textbf{PE:} \foreignlanguage{vietnamese}{\colorbox{lightorange}{Du lịch bằng đường thủy} \colorbox{lightblue}{nội địa} là một \colorbox{lightorange}{ý tưởng nghỉ dưỡng không tồi.}} \\
        \bottomrule
    \end{tabular}
    \caption{An example sentence (81.1) from the DivEMT corpus, with the English source and all output modalities for all target languages, including intermediate machine translations (MT) and subsequent post-editings (PE). Colors denote \colorbox{lightgreen}{insertions}, \colorbox{lightred}{deletions}, \colorbox{lightorange}{substitutions} and \colorbox{lightblue}{shifts} computed with Tercom~\cite{snover-etal-2006-study}.}
    \label{tab:data_example_full_1}
\end{table*}

\begin{table*}
    \centering
    \footnotesize
    \begin{tabular}{p{0.2in} p{\textwidth - 0.6in}}
        \toprule
        \textsc{English} \\
        & \vspace{0.3mm} The Internet combines elements of both mass and interpersonal communication. \\
        \midrule
        \textsc{Arabic} \\ \vspace{0.2mm}
        HT & \vspace{-1.5mm} \setcode{utf8}{\< يجمع الإنترنت بين عناصر وسائل الاتصال العامة والشخصية على حدٍ سواء. >}\\
        \cmidrule(lr){2-2}
        \multirow{2}{*}{PE$_1$} 
        & \textbf{MT:} \includegraphics[height=5mm]{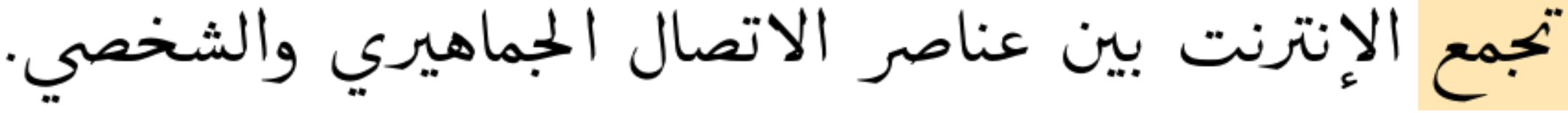} \\% \< تجمع الإنترنت بين عناصر الاتصال الجماهيري والشخصي. >  \\%
        & \textbf{PE:} \includegraphics[height=5mm]{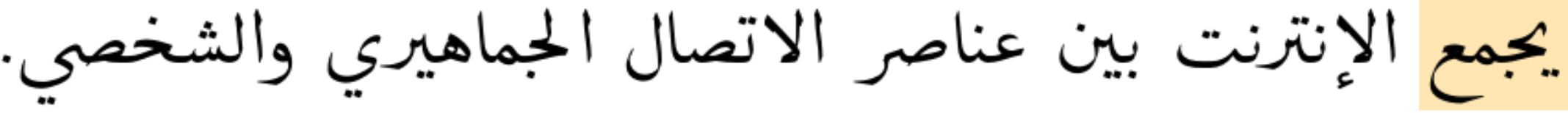} \\% \< يجمع الإنترنت بين عناصر الاتصال الجماهيري والشخصي. >  \\%
        \cmidrule(lr){2-2}
        \multirow{2}{*}{PE$_2$} 
        & \textbf{MT:} \includegraphics[height=5mm]{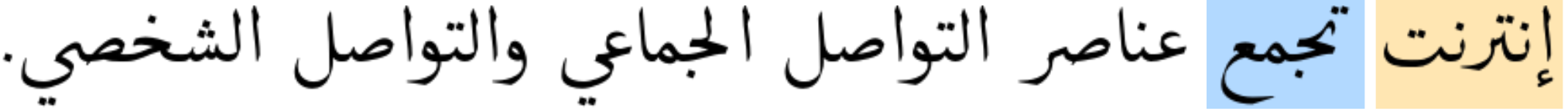} \\% \< إنترنت تجمع عناصر التواصل الجماعي والتواصل الشخصي. >  \\%
        & \textbf{PE:} \includegraphics[height=5mm]{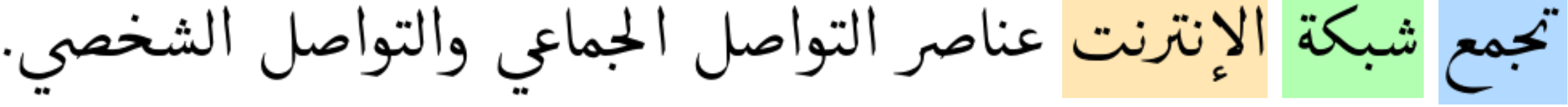} \\% \< تجمع شبكة الإنترنت عناصر التواصل الجماعي والتواصل الشخصي. >  \\%
        \midrule
        \textsc{Dutch} \\ \vspace{0.3mm}
        HT & \vspace{0.3mm} Het internet combineert elementen van zowel massa- en intermenselijke communicatie. \\
        \cmidrule(lr){2-2}
        \multirow{2}{*}{PE$_1$} & \textbf{MT:} Het internet combineert elementen van zowel massa- als interpersoonlijke communicatie. \vspace{1.3mm}\\
        & \textbf{PE:} Het internet combineert elementen van zowel massa- als interpersoonlijke communicatie. \\
        \cmidrule(lr){2-2}
        \multirow{2}{*}{PE$_2$} & \textbf{MT:} Het internet combineert elementen van massa- en interpersoonlijke communicatie. \vspace{1.3mm}\\
        & \textbf{PE:} Het internet combineert elementen van massa- en interpersoonlijke communicatie. \\
        \midrule
        \textsc{Italian} \\
        \vspace{0.3mm}
        HT & \vspace{0.3mm} Internet combina elementi di comunicazione sia di massa sia interpersonale. \\
        \cmidrule(lr){2-2}
        \multirow{2}{*}{PE$_1$} & \textbf{MT:} Internet combina elementi di comunicazione di massa e \colorbox{lightorange}{interpersonali}. \\
        & \textbf{PE:} Internet combina elementi di comunicazione di massa e \colorbox{lightorange}{interpersonale}. \\
        \cmidrule(lr){2-2}
        \multirow{2}{*}{PE$_2$} & \textbf{MT:} Internet combina elementi di comunicazione di massa e interpersonale. \vspace{1.3mm}\\
        & \textbf{PE:} Internet combina elementi di comunicazione di massa e interpersonale. \\
        \midrule
        \textsc{Turkish} \\ \vspace{0.3mm}
        HT & \vspace{0.3mm} İnternet hem kitlesel hem de bireysel iletişim öğelerini birleştiriyor. \\
        \cmidrule(lr){2-2}
        \multirow{2}{*}{PE$_1$} 
        & \textbf{MT:} İnternet, hem \colorbox{lightorange}{kitle} hem de kişiler arası iletişimin unsurlarını birleştirir.\\
        & \textbf{PE:} İnternet, hem \colorbox{lightorange}{kitleler} hem de kişiler arası iletişimin unsurlarını birleştirir.\\
        \cmidrule(lr){2-2}
        \multirow{2}{*}{PE$_2$}
        & \textbf{MT:} İnternet hem kitlesel hem de kişisel iletişim unsurlarını birleştiriyor. \vspace{1.3mm}\\
        & \textbf{PE:} İnternet hem kitlesel hem de kişisel iletişim unsurlarını birleştiriyor. \\
        \midrule
        \textsc{Ukrainian} \\ \vspace{0.3mm}
        HT & \vspace{0.3mm} \foreignlanguage{ukrainian}{В інтернеті поєднуються елементи групового спілкування та особистого спілкування.} \\
        \cmidrule(lr){2-2}
        \multirow{2}{*}{PE$_1$} 
        & \textbf{MT:} \foreignlanguage{ukrainian}{Інтернет поєднує в собі елементи як масового, так і міжособистісного спілкування. \vspace{1.3mm}} \\
        & \textbf{PE:} \foreignlanguage{ukrainian}{Інтернет поєднує в собі елементи як масового, так і міжособистісного спілкування.} \\
        \cmidrule(lr){2-2}
        \multirow{2}{*}{PE$_2$}
        & \textbf{MT:} \foreignlanguage{ukrainian}{Інтернет \colorbox{lightorange}{об 'єднує} як \colorbox{lightorange}{масову}, так і \colorbox{lightorange}{міжлюдську комунікацію.}} \\
        & \textbf{PE:} \foreignlanguage{ukrainian}{Інтернет \colorbox{lightgreen}{поєднує в} \colorbox{lightorange}{собі елементи} як \colorbox{lightorange}{групової}, так і \colorbox{lightorange}{особистої комунікації.}} \\
        \midrule
        \textsc{Vietnamese} \\ \vspace{0.3mm}
        HT & \vspace{0.3mm}
        \foreignlanguage{vietnamese}{Internet là nơi tổng hợp các yếu tố của cả phương tiện truyền thông đại chúng và giao tiếp liên cá nhân.} \\
        \cmidrule(lr){2-2}
        \multirow{2}{*}{PE$_1$} 
        & \textbf{MT:} \foreignlanguage{vietnamese}{Internet kết hợp các yếu tố của cả \colorbox{lightorange}{giao tiếp} đại chúng và giao tiếp giữa các cá nhân.} \\
        & \textbf{PE:} \foreignlanguage{vietnamese}{Internet kết hợp các yếu tố của cả \colorbox{lightorange}{truyền thông} đại chúng và giao tiếp giữa các cá nhân.} \\
        \cmidrule(lr){2-2}
        \multirow{2}{*}{PE$_2$}
        & \textbf{MT:} \foreignlanguage{vietnamese}{Internet kết hợp những yếu tố của \colorbox{lightorange}{sự} giao tiếp \colorbox{lightorange}{quần} chúng và giao tiếp \colorbox{lightred}{giữa người} \colorbox{lightorange}{với người}.} \\
        & \textbf{PE:} \foreignlanguage{vietnamese}{Internet kết hợp những yếu tố của \colorbox{lightgreen}{cả} \colorbox{lightorange}{việc} giao tiếp \colorbox{lightorange}{đại} chúng và giao tiếp \colorbox{lightorange}{cá nhân}.} \\
        \bottomrule
    \end{tabular}
    \caption{An example sentence (29.2) from the DivEMT corpus, with the English source and all output modalities for all target languages, including intermediate machine translations (MT) and subsequent post-editings (PE). Colors denote \colorbox{lightgreen}{insertions}, \colorbox{lightred}{deletions}, \colorbox{lightorange}{substitutions} and \colorbox{lightblue}{shifts} computed with Tercom~\cite{snover-etal-2006-study}.}
    \label{tab:data_example_full_2}
\end{table*}

\section{Model Description and Feature Significance}
\label{app:modeling}

\begin{figure}[t]
    \centering
    \includegraphics[width=\linewidth]{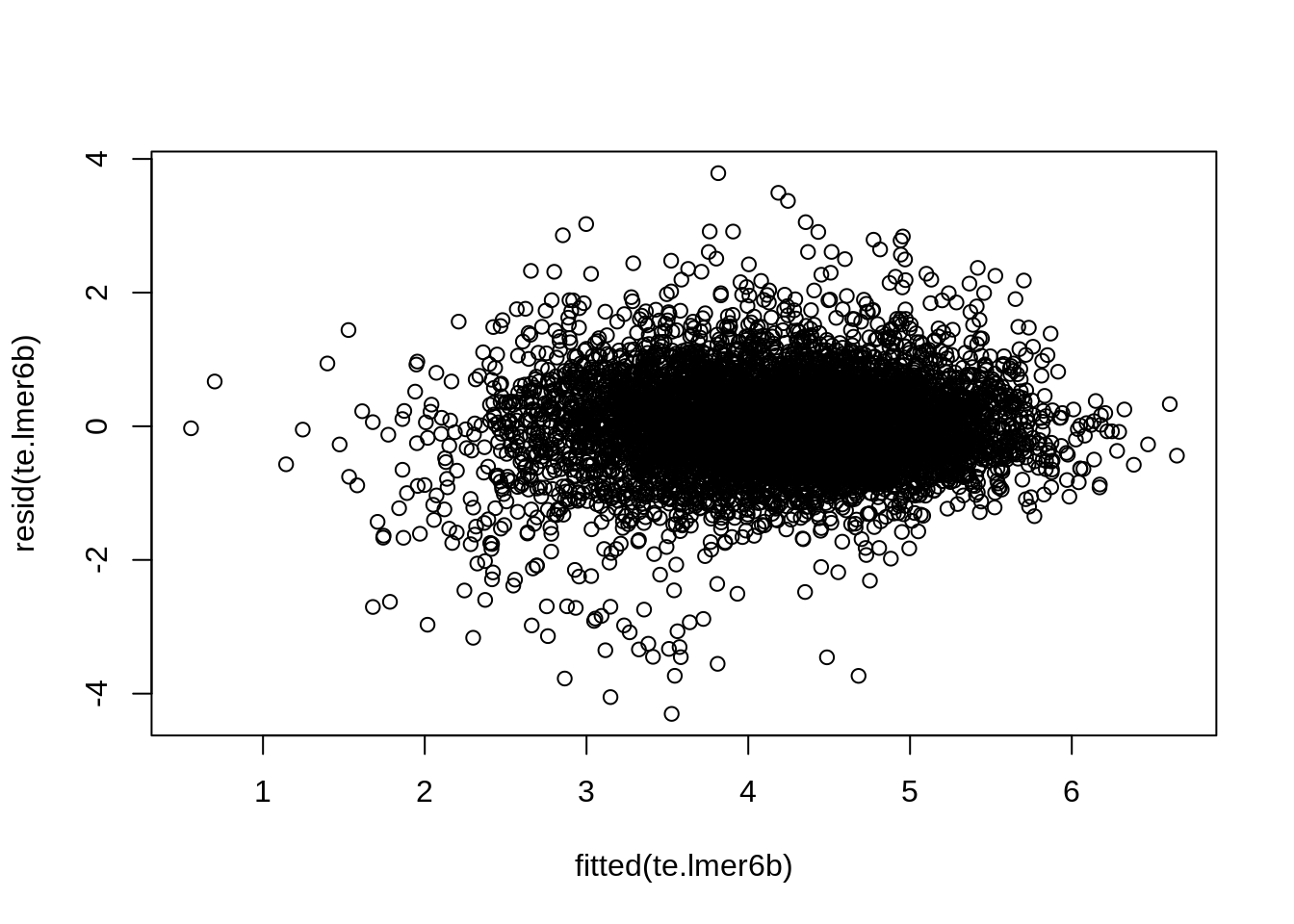}
  \caption{Residuals of the final LMER model, used to verify the heteroscedasticity assumption.}
  \label{fig:residuals}
\end{figure}

\begin{figure}[t]
    \centering
    \includegraphics[width=\linewidth]{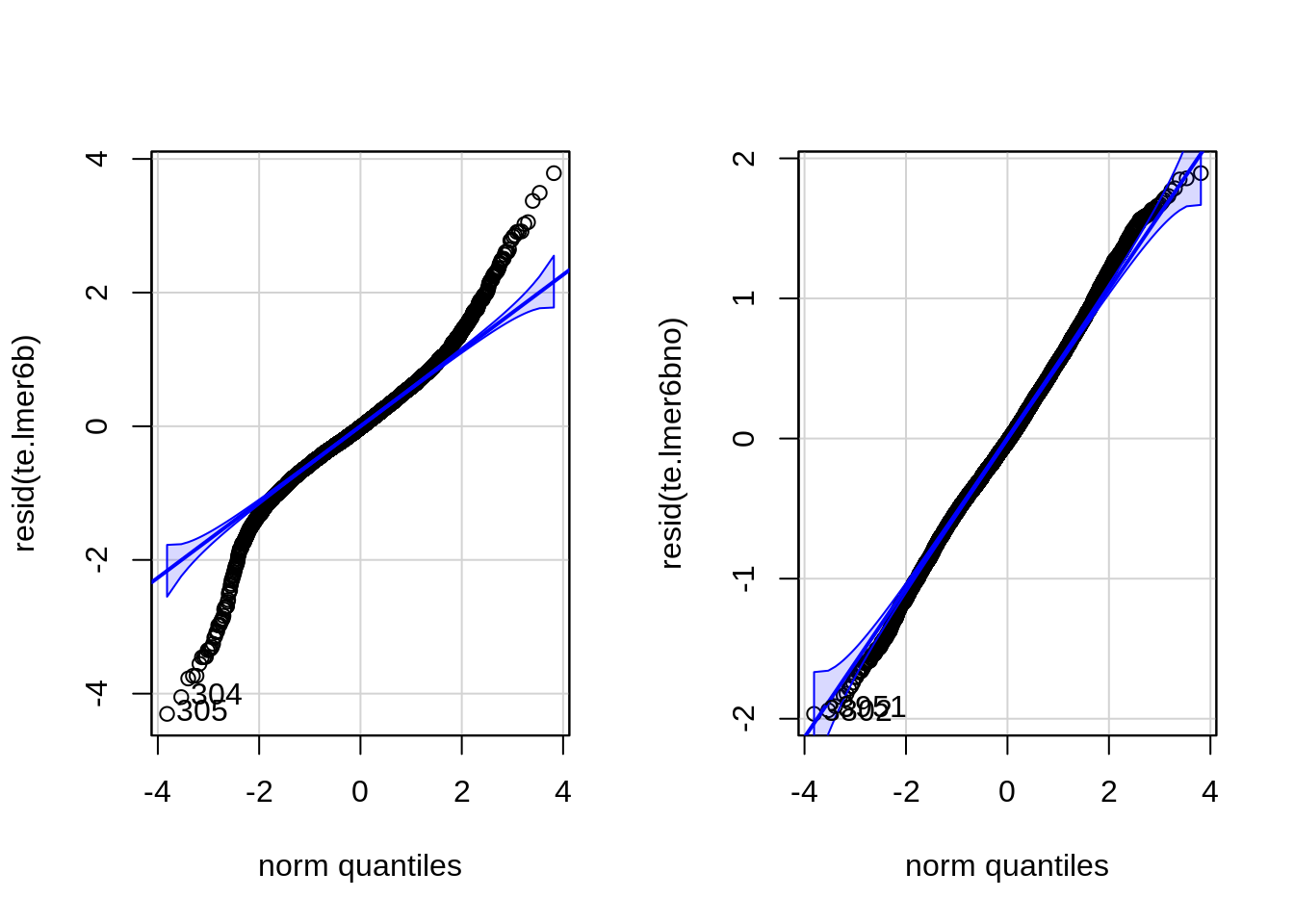}
  \caption{Quantile-quantile plot before and after the removal of outliers when fitting the LMER model, used to verify the normality assumption.}
  \label{fig:normality}
\end{figure}

\begin{table}
    \small
    \centering
    \begin{tabular}{lr}
        \toprule
         \textbf{Subject} & \textbf{Coefficient} \\
        \midrule
        \texttt{ara\_t1} & 0.281  \\
        \texttt{ara\_t2} & -0.384  \\
        \texttt{ara\_t3} & -0.103  \\
        \midrule
        \texttt{nld\_t1} & 0.001  \\
        \texttt{nld\_t2} & -0.459  \\
        \texttt{nld\_t3} & 0.458  \\
        \midrule
        \texttt{ita\_t1} & 0.086  \\
        \texttt{ita\_t4} & 0.350  \\
        \texttt{ita\_t5} & -0.436  \\
        \midrule
        \texttt{tur\_t1} & -0.381  \\
        \texttt{tur\_t2} & 0.272  \\
        \texttt{tur\_t3} & 0.109  \\
        \midrule
        \texttt{ukr\_t1} & 0.077  \\
        \texttt{ukr\_t2} & 0.314  \\
        \texttt{ukr\_t3} & -0.391  \\
        \midrule
        \texttt{vie\_t1} & 0.012  \\
        \texttt{vie\_t2} & 0.176  \\
        \texttt{vie\_t3} & -0.188  \\
        \bottomrule
    \end{tabular}
    \caption{Coefficients of the random intercept related to the \texttt{subject\_id} variable, representing the identity of the translator performing the translation.}
  \label{tab:coeff-reff-lmer}
\end{table}

Linear Mixed Effects models (LMER) are used for regression analyses involving dependent data, such as longitudinal studies with multiple observations per subject. Given the variables of Table~\ref{tab:divemt-fields}, our final model to predict translation time has the following formulation:
$$\begin{aligned}
   & \small\texttt{edit\_time} \sim \;\small\texttt{src\_len\_chr + lang\_id * task\_type} \\
   & \small\texttt{+ (1|subject\_id)} \\
   & \small\texttt{+ (1 | document\_id/item\_id)} \\
   & \small\texttt{+ (0 + task\_type | document\_id/item\_id)} \\
\end{aligned}$$
We log-transform the dependent variable, edit time in seconds, given its long right tail. The models are built by adding one element at a time, and checking whether such addition leads to a significantly better model with AIC (i.e. if the score gets reduced by at least 2). We fit the models using ML when comparing models that differ in the fixed structure, and REML when they differ in the random structure.
We start with an initial model that just includes the two random intercepts (by-translator and by-segment) and proceed by (i) finding significance for nested document/segment random effect; (ii) adding fixed predictors one by one; (iii) adding interactions between fixed predictors; and (iv) adding the random slopes.

From this sequential procedure, we obtain the resulting model. When checking the homoscedasticity and normality of residuals assumptions (Figures~\ref{fig:residuals} and~\ref{fig:normality}), we find the latter is not fulfilled. Consequently, we remove data points for which observations deviate by more than 2.5 standard deviations from the predicted value by the model (2.4\% of the data) and refit the best model on this subset, in order to find out whether any of the effects were due to these outliers. The resulting trends do not change significantly in this final model, in which residuals are normally distributed. As a final sanity check, in Table~\ref{tab:coeff-reff-lmer} we measure the effect of subject identity on edit times and find no systematic patterns across languages.

\end{document}